\documentclass[10pt,twocolumn,letterpaper]{article}

\usepackage{cvpr}              

\definecolor{cvprblue}{rgb}{0.21,0.49,0.74}
\usepackage[pagebackref,breaklinks,colorlinks,allcolors=cvprblue]{hyperref}

\usepackage[table]{xcolor} 
\usepackage{algorithm}
\usepackage{algorithmic}
\usepackage{multirow, makecell}
\usepackage{amsmath}
\usepackage{amsfonts}
\usepackage{capt-of,etoolbox}
\usepackage[dvipsnames]{xcolor}
\usepackage{graphicx}
\usepackage[capitalize]{cleveref}
\usepackage{subcaption}
\usepackage{enumitem}
\usepackage{pifont}
\usepackage{amssymb}
\usepackage{nicefrac}       
\usepackage{booktabs}       
\usepackage{amsfonts}       
\usepackage[accsupp]{axessibility}  

\makeatletter
\@namedef{ver@eso-pic.sty}{9999/12/31}
\makeatother
\usepackage{pdfpages}
\usepackage{setspace}

\newtheorem{proposition}{Proposition}

\def\ie{\emph{i.e.}}

\def\wrt{w.r.t.}

\newcommand{\x}[1]{\boldsymbol{x}_{#1}}

\newcommand{\xtilde}[1]{\tilde{\boldsymbol{x}}_{#1}}
\newcommand{\driftConst}[1]{f({#1})}
\newcommand{\diffConst}[1]{g({#1})}
\newcommand{\score}[2]{\nabla_{\x{t}} \log {#1}({#2})}

\newcommand{\imgres}[1]{$({#1} \times {#1})$}

\newcommand{\alp}[1]{\alpha_{#1}}

\newcommand{\sig}[1]{\sigma_{#1}}
\newcommand{\lam}[1]{\lambda_{#1}}

\newcommand{\dif}[1]{\text{d}{#1}}
\def\mathbi#1{\textbf{\em #1}}

\newcommand{\ddt}[1]{\frac{\text{d} {#1}}{\text{d}t}}

\def\paperID{4583} 
\def\confName{CVPR}
\def\confYear{2026}

\title{DBMSolver: A Training-free Diffusion Bridge Sampler\\for High-Quality Image-to-Image Translation}

\author{Sankarshana Venugopal\\\\
{\tt\small sankarshana.v@gmail.com}
\and
Mohammad Mostafavi\\
Seoul National University\\
{\tt\small mostafavi.isfahani@gmail.com}
\and
Jonghyun Choi\thanks{Corresponding author. JC is with ECE, IPAI, ASRI in SNU.}\\\\
{\tt\small jonghyunchoi@snu.ac.kr
}
}

\newcommand\myfigure{
  \vspace{-2.5em}
  \begin{center}
    \includegraphics[width=0.8\textwidth]{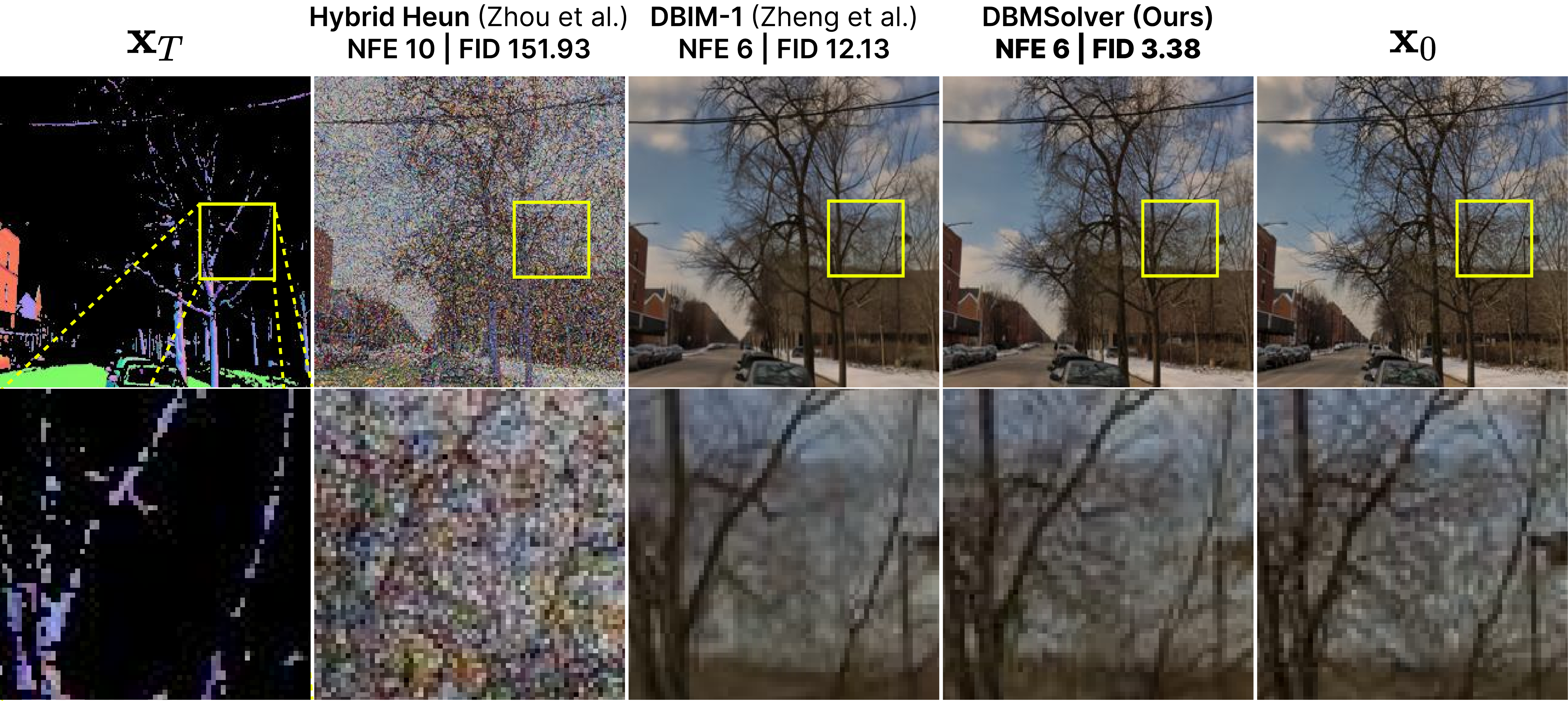}
    \captionof{figure}{Few-step image synthesis (6 NFEs $\downarrow$) with high-quality generated details (3.38 FID $\downarrow$) on DIODE \imgres{256}~\cite{diode}.}
    \label{fig:teaser}
  \end{center}
  \vspace{1em}
}

\begin{document}

\makeatletter
\apptocmd\@maketitle{{\myfigure{}\par}}{}{}
\makeatother

\maketitle

\begin{abstract}
Diffusion-based image-to-image (I2I) translation excels in high-fidelity generation but suffers from slow sampling in state-of-the-art Diffusion Bridge Models (DBMs), often requiring dozens of function evaluations (NFEs). We introduce \textbf{DBMSolver}, a training-free sampler that exploits the semi-linear structure of DBM's underlying SDE and ODE via exponential integrators, yielding highly-efficient $1^\text{st}$- and $2^\text{nd}$-order solutions. This reduces NFEs by up to $5\times$ while boosting quality (e.g., FID drops $53\%$ on DIODE at 20 NFEs vs. $2^\text{nd}$-order baseline). Experiments on inpainting, stylization, and semantics-to-image tasks across resolutions up to 256$\times$256 show DBMSolver sets new SOTA efficiency-quality tradeoffs, enabling real-world applicability. Our code is publicly available at \href{https://github.com/snumprlab/dbmsolver}{https://github.com/snumprlab/dbmsolver}.
\end{abstract}
\vspace{-1.5em}

\section{Introduction}
\label{sec:intro}

Image-to-Image (I2I) Translation is a generative modeling paradigm that learns to map an input image to a target output. It encompasses tasks like image restoration, grayscale colorization, and inpainting of occluded or corrupted regions, as well as style transfer and semantic reinterpretation via cross-domain synthesis~\cite{face2comics,pix2pix,goodfellow2014generative}.

Recent diffusion-based works, as alternatives to traditional generative approaches such as GANs~\cite{cyclegan, stylegan2}, have brought significant advances in the synthesis of high-fidelity images~\cite{palette,i2sb,ddrm}. 
Among them,~\cite{ddbm} proposed Diffusion Bridge Models (DBMs), capable of performing I2I Translation by establishing a \textit{diffusion bridge} that facilitates the translation from one arbitrary image distribution to another.

While DBMs offer a theoretically elegant diffusion framework for I2I Translation, generating high-quality images using such diffusion-based models remains computationally expensive, as it requires numerous costly function evaluations (NFEs)--which we aim to tackle in this work \textbf{without requiring any additional training overhead}.

\Cref{fig:teaser} presents the fact that while Hybrid Heun~\cite{ddbm} and DBIM~\cite{dbim} require numerous NFEs for coherent or high-quality outputs on DIODE~\cite{diode}, our method (DBMSolver) achieves superior fidelity in just 6 NFEs, preserving intricate details like tree branches without artifacts.

Unlike DBIM's linear multistep methods~\cite{dbim} or audio-specific EI in Bridge-TTS~\cite{btts}, DBMSolver derives closed-form solutions for VP/VE bridges, avoiding crude approximations and enabling FID gains even at low NFEs.

\subsection{DBMSolver Overview}
\label{subsec:intro-dbmsolver}

We introduce \textit{DBMSolver}: a training-free, highly efficient solver specifically designed to accelerate the DBM-based sampling process. 
Previous research on improving the sampling speeds explored either model distillation~\cite{cdbm, em_dist, invbridge}, fine-tuning~\cite{cmsmadeeasy, cdbm}, or re-training of an entire neural network~\cite{imm}.
In contrast, DBMSolver is a drop-in replacement for existing DBM sampling methods, avoiding the need for any architectural changes or extra training, thus enabling broad compatibility and immediate benefits.

We devise DBMSolver by rigorously analyzing the underlying Stochastic and Ordinary Differential Equations (SDE and ODE) governing DBMs' reverse-time diffusion process~(Equations~\ref{eq:ddbm_backward_sde} and~\ref{eq:ddbm_backward_ode}).
Specifically, in~\Cref{subsec:sde_soln}, we identify the inherent semi-linear structure of the SDE and leverage the Exponential Integrators (EI)~\cite{expint} method to derive an efficient $1^\text{st}$-order solution.
Next, in~\Cref{subsec:ode_soln}, we analyze the ODE to show that it has a semi-linear structure as well, allowing us to formulate an exact solution that we efficiently approximate with a $2^\text{nd}$-order solution.
By combining the two ODE/SDE solutions, we devise a sampling procedure that drastically reduces the required NFEs while  enhancing the image quality, as showcased in~\Cref{subsec:devising_dbmsolver}.
The key contributions of this work are as follows:
\begin{itemize}
    \item \textbf{A novel, training-free bridge sampler}: DBMSolver accelerates DBM sampling by 20x (e.g., 6 NFEs vs. 119 for Hybrid Heun) on conditional/unconditional I2I tasks, with zero retraining--unlike distillation~\cite{cdbm} or fine-tuning~\cite{invbridge} approaches.
    \item \textbf{Theoretically grounded}: We provide analytical solutions for the Bridge SDE and ODE governing DBMs, grounded in diffusion bridge theory.
    \item \textbf{State-of-the-art performances on various I2I tasks}: Through extensive experimentation on various I2I tasks and image resolutions, we show that DBMSolver consistently achieves state-of-the-art results, outperforming prior arts~\cite{dbim, bbdm, odes3} in terms of generation quality and computational efficiency.
\end{itemize}

\section{Preliminaries and Related Work}
\label{sec:prelims}

\begin{table*}[ht!]
\centering
\caption{\textbf{Comparison of diffusion samplers}. N2I methods assume Gaussian priors, invalid for DBMs' arbitrary $p_T(\mathbf{x})$. Our DBMSolver uses Exponential Integrator (EI)~\cite{expint} solutions for bridges, enabling Markovian, higher-order sampling.}
\vspace{-0.5em}
\resizebox{\linewidth}{!}{
\begin{tabular}{lccccc}
\toprule
\multirow{2}{*}{Sampling Method} & \multirowcell{2}{Assumption of\\Prior Distribution $p_T(\x{})$} & \multirowcell{2}{Theoretically valid on\\Image-to-Image Translation} & \multirow{2}{*}{Order} & \multirow{2}{*}{Sampling Procedure} & \multirow{2}{*}{Is Markovian?} \\
&&&\\
\midrule
\rowcolor{gray!10} 
\multicolumn{6}{l}{\textit{Samplers designed for N2I-based DPMs:}}\\
\midrule
DDIM~\cite{ddim} & $\x{T} \sim \mathcal{N}(\mathbf{0}, \sig{T}^2\mathbi{I})$ & {\color{red}\ding{55}} & $1^\text{st}$ & $p_t(\x{t_{i-1}} \mid \x{t_i})$ & {\color{red}\ding{55}} \\
\midrule
DPMSolver++(2M)~\cite{dpmpp} & $\x{T} \sim \mathcal{N}(\mathbf{0}, \sig{T}^2\mathbi{I})$ & {\color{red}\ding{55}} & $2^\text{nd}$ & Analytic Soln. of ODE via EI~\cite{expint} & {\color{ForestGreen}\ding{51}} \\
\midrule
\rowcolor{gray!10} 
\multicolumn{6}{l}{\textit{Samplers designed for I2I-based DBMs:}}\\
\midrule
Hybrid Heun (HH)~\cite{ddbm} & $\x{T} \sim p_\text{prior}(\x{})$ & {\color{ForestGreen}\ding{51}} & $2^\text{nd}$ & Alternating Bridge SDE (Euler-Maruyama) \& ODE (Heun) Steps & {\color{ForestGreen}\ding{51}} \\
\midrule
ODES3 \cite{odes3} & $\x{T} \sim p_\text{prior}(\x{})$ & {\color{ForestGreen}\ding{51}} & $2^\text{nd}$ & Initial Bridge SDE (Euler-Maruyama) with Subsequent ODE (Heun) Steps & {\color{ForestGreen}\ding{51}} \\
\midrule
DBIM-1~\cite{dbim} & $\x{T} \sim p_\text{prior}(\x{})$ & {\color{ForestGreen}\ding{51}} & $1^\text{st}$ & $p_t(\x{t_{i-1}} \mid \x{t_i}, \x{T})$ & {\color{red}\ding{55}} \\
\midrule
\multirow{3}{0.15\linewidth}{DBIM-2~\cite{dbim}\\and\\DBIM-3~\cite{dbim}} & \multirow{3}{0.15\linewidth}{\centering $\x{T} \sim p_\text{prior}(\x{})$} & \multirow{3}{0.1\linewidth}{\centering{\color{ForestGreen}\ding{51}}} & \multirow{3}{0.1\linewidth}{\centering $2^\text{nd}$\\and\\$3^\text{rd}$}  & \multirow{3}{0.4\linewidth}{\centering Analytic Soln. of Bridge SDE \& ODE via\\EI~\cite{expint}, with Linear Multistep methods\\(numerical; not closed-form)} & \multirow{3}{0.15\linewidth}{\centering {\color{Red}\ding{55}}} \\ \\
\\
\midrule
\rowcolor{gray!20}
DBMSolver \textbf{(Ours)} & $\x{T} \sim p_\text{prior}(\x{})$ & {\color{ForestGreen}\ding{51}} & $2^\text{nd}$ & Analytic Soln. of Bridge SDE \& ODE via EI~\cite{expint} & {\color{ForestGreen}\ding{51}} \\
\bottomrule
\end{tabular}
}
\vspace{2pt}
\label{tab:methodology_comp_table}
\vspace{-1em}
\end{table*}

\subsection{Diffusion-based Generative Models}
\label{subsec:diffusion_models}

\paragraph{Diffusion Probabilistic Models (DPMs).}
Owing to their ability to generate high-quality outputs, DPMs have become ubiquitous for various noise-to-image generation tasks~\cite{ldm, karrasedm}.
DPMs learn to traverse from a Gaussian distribution $p_\text{prior}(\mathbf{x})$ to an unknown data distribution $p_0(\mathbf{x}) := p_\text{data}(\mathbf{x})$ through a gradual denoising process~\cite{ddpm, songscore, dmdhariwal}.
In other words, starting from a prior distribution $p_T(\mathbf{x}) := p_\text{prior}(\mathbf{x}) \approx \mathcal{N}(\mathbf{0}, \sigma_T^2\,\mathbi{I})$ with $\sig{T} > 0$, DPMs iteratively denoise $\x{T} \sim p_T(\mathbf{x})$ (\ie, white noise) to recover the desired output $\x{0} \sim p_0(\mathbf{x})$.
This reverse diffusion process is shown to follow the Ordinary Differential Equation (ODE)~\cite{anderson, songscore}:
\begin{equation}
    \dif{\x{t}}=\left[ \, \driftConst{\x{t}, t}-\frac{1}{2} \diffConst{t}^2 \score{p_t}{\x{}} \right] \dif{t},
\label{eq:backward_ode}
\end{equation}
where $p_t(\mathbf{x})$ is the marginal distribution of $\x{t}$ at $t$, and $\score{p_t}{\mathbf{x}}$ is its \textit{score function} learned by a neural network~\cite{hyvarinen05a}, and $\,\driftConst{\x{t}, t}$ and $\diffConst{t}$ are the drift and diffusion coefficients, respectively~(see Supplementary).
\cite{songscore} terms this the \textit{Probability Flow (PF) ODE}.

\vspace{-1em}
\paragraph{Diffusion Bridge Models (DBMs).}
Although DPMs have gained popularity for N2I Generation tasks, their underlying theory only holds when the prior distribution is purely Gaussian, \ie, $p_T(\mathbf{x}) \approx \mathcal{N}(\mathbf{0}, \sigma_T^2\,\mathbi{I})$.
However, this assumption does not hold for I2I translation tasks where $p_T(\mathbf{x})$ is not necessarily Gaussian noise, leading to output images that do not remain faithful to the original $\x{0}$, limiting their applicability in such settings.

To solve this,~\cite{ddbm} extend the diffusion framework from N2I Generation to I2I Translation by making use of Doob's h-transform~\cite{doob, rogers}.
By steering the forward diffusion process almost surely to a target via Doob's h-transform, they form a \textit{diffusion bridge} between $\x{0} \sim p_0(\mathbf{x})$ and $\x{T} \sim p_T(\mathbf{x})$, yielding a \textit{conditioned} forward diffusion process.

The corresponding reverse-time process is governed by the Bridge SDE:
\begin{align}
    \dif{\x{t}} & = \driftConst{\x{t}, t} \dif{t}  - \diffConst{t}^2 \score{p_t}{\x{t} \mid \x{T}} \dif{t} \notag \\
    & + \diffConst{t}^2\score{p_t}{\x{T} \mid \x{t}} \dif{t} + \diffConst{t} \, \dif{\mathbf{w}_t},
\label{eq:ddbm_backward_sde}
\end{align}
where $\score{p_t}{\x{t} \mid \x{T}}$ is the score of the tractable \textit{conditional probability}, $p_t(\x{t} \mid  \x{T})$:
\begin{align}
\frac{\frac{\text{SNR}_T}{\text{SNR}_t} \frac{\alp{t}}{\alp{T}} \x{T} + \alp{t} \left( 1 - \frac{\text{SNR}_T}{\text{SNR}_t} \right) \x{0} - \x{t}}{\sig{t}^2 \left( 1 - \frac{\text{SNR}_T}{\text{SNR}_t} \right)},
\label{eq:score_leanred_by_dbm}
\end{align}
which is learned by a DBM via Bridge Score Matching~\cite{ddbm} (\ie, $\mathbi{s}_{\boldsymbol{\theta}}(\x{t}, t, \x{T}) \approx \score{p_t}{\x{t} \mid \x{T}}$).
The score of the \textit{transition probability}, $p_t(\x{T} \mid \x{t})$, is:
\begin{align}
\score{p_t}{\x{T} \mid \x{t}} = \frac{\frac{\alp{t}}{\alp{T}} \x{T} - \x{t}}{\sig{t}^2 \left( \frac{\text{SNR}_t}{\text{SNR}_T} - 1 \right)}, \, \text{SNR}_t := \nicefrac{\alp{t}^2}{\sig{t}^2},
\label{eq:score_of_h_function}
\end{align}
where SNR$_t$ is the signal-to-noise ratio at time $t$.
Lastly, the SDE in~\Cref{eq:ddbm_backward_sde} has an equivalent ODE interpretation, which we name \emph{Bridge Probability Flow (PF) ODE}:
\vspace{-1em}
\begin{align}
    \ddt{\x{t}} = & \, \driftConst{\x{t}, t} - \frac{1}{2} \diffConst{t}^2  \score{p_t}{\x{t} \mid \x{T}} \notag \\
    & + \diffConst{t}^2  \score{p_t}{\x{T} \mid \x{t}}.
\label{eq:ddbm_backward_ode}
\end{align}

\vspace{-1em}
\subsection{Fast Samplers for DMs and DBMs}
\label{subsec:fast_samplers_for_dms}
For DPM-based N2I Generation, works such as~\cite{dpm, dpmpp, unipc} proposed fast samplers that generate high-quality images in $\leq 20$ NFEs.
These methods follow the assumption that the prior is a pure Gaussian distribution.
However, since this assumption becomes \textbf{invalid} for I2I Translation (as prior $p_T(\mathbf{x})$ can be arbitrary), their theoretical foundation is unsuitable for I2I tasks, calling for samplers that support arbitrary priors.
Table~\ref{tab:methodology_comp_table} contrasts samplers, highlighting why N2I methods (e.g., DDIM~\cite{ddim}) fail on DBMs: their Gaussian prior assumption breaks when $p_T(\mathbf{x})$ is not Gaussian, leading to an invalid bridge (see Supp. for examples).

To generate high-quality images with DBMs,~\cite{ddbm} proposed the Hybrid Heun (HH) Sampler, which alternatively solves the Bridge SDE (\Cref{eq:ddbm_backward_sde}) via $1^\text{st}$-order Euler-Maruyama and the Bridge PF ODE (\Cref{eq:ddbm_backward_ode}) via $2^\text{nd}$-order Heun.
Next, inspired by~\cite{ddim},~\cite{dbim} proposed a non-Markovian $1^\text{st}$-order sampler called DBIM (``DBIM-1''). 
``DBIM-2" and ``DBIM-3" have also been proposed, derived via Linear Multistep methods analogous to DPMSolver++(2M)~\cite{dpmpp}, serving as $2^\text{nd}$- and $3^\text{rd}$-order samplers, respectively.
A recent finding, ODES3~\cite{odes3}, implements a straightforward algorithm with $1^\text{st}$-order Euler-Maruyama initial step with the Bridge SDE and subsequent $2^\text{nd}$-order Heun steps with the Bridge ODE.

In contrast, we analyze and rigorously derive exact solutions for the Bridge SDE and ODE to propose a highly-efficient $2^\text{nd}$-order sampler that surpasses prior works in image quality and efficiency.
DBMSolver is capable of handling arbitrary priors while achieving better results compared to previous HH, DBIM-2/3, and ODES3 samplers.

\section{DBMSolver: A Fast DBM Sampler}
\label{sec:dbmsolver}
DBMs' diffusion bridge introduces non-Gaussian drifts that N2I solvers do not take into account. We uncover their overlooked semi-linear form--linear in $\mathbf{x}_t$ with non-linear scores--and provide exact linear-term solutions that prior works missed.
The fundamental difference between DPMs and DBMs is that $\x{T}$ is pure noise for DPMs, but it can be an arbitrary image for DBMs.
Consequently, DBMs involve a reverse diffusion process conditioned on $\x{T}$, which is crucial for I2I Translation.
As discussed in~\Cref{subsec:fast_samplers_for_dms}, this crucial distinction invalidates the direct application of state-of-the-art fast N2I solvers~\cite{dpmpp} for sampling DBMs.
We explore a different approach to developing fast samplers specifically for DBMs by thoroughly analyzing their underlying reverse diffusion SDE and ODE (Eqs. \ref{eq:ddbm_backward_sde} \&~\ref{eq:ddbm_backward_ode}), and deriving analytic solutions that eliminate approximation errors associated with the linear terms using Exponential Integrators (EI)~\cite{expint}.
With these solutions, we develop a higher-order sampling procedure that generates high-quality images significantly faster, tailored for DBMs.

Let $\mathbi{D}_{\boldsymbol{\theta}}(\x{s}, s, \x{T}, T)$ denote an $\x{0}$-predicting DBM such that $\mathbi{D}_{\boldsymbol{\theta}}(\x{s}, s, \x{T}, T) \approx \x{0}$ for $s \in [0, T]$. For brevity, we adopt $\mathbi{D}_{\boldsymbol{\theta}}(\x{s}) := \mathbi{D}_{\boldsymbol{\theta}}(\x{s}, s, \x{T}, T)$.

\subsection{Uncovering the Semi-Linear Structures}
To derive an efficient solver, we first reformulate the reverse process. 
From the definition of $p_t(\x{t} \mid \x{T})$ in~\Cref{eq:score_leanred_by_dbm}, its score and the $\x{0}$-predicting DBM $\boldsymbol{D_\theta}$ are related via:
\begin{equation}
    \mathbi{s}_{\boldsymbol{\theta}}(\x{t}) = \frac{ \frac{\text{SNR}_T}{\text{SNR}_t} \frac{\alp{t}}{\alp{T}} \x{T} + \alp{t} \left( 1 - \frac{\text{SNR}_T}{\text{SNR}_t} \right) \mathbi{D}_{\boldsymbol{\theta}}(\x{t}) - \x{t}}{\sig{t}^2 \left( 1 - \frac{\text{SNR}_T}{\text{SNR}_t} \right)},
    \label{eq:score_to_x0}
\end{equation}
where $\mathbi{s}_{\boldsymbol{\theta}}(\x{t}) \approx \score{p_t}{\x{t} \mid \x{T}}$.
By substituting the $\x{0}$-predictor $\mathbi{D}_{\boldsymbol{\theta}}$ back into the Bridge PF ODE (Eq. \ref{eq:ddbm_backward_ode}), the ODE can be rewritten as a \textit{semi-linear} equation:
\begin{equation}
    \ddt{\x{t}} = \underbrace{L(t) \, \x{t}}_{\text{Linear term}} + \underbrace{N(\mathbi{D}_{\boldsymbol{\theta}}(\x{t}), t, \x{T})}_{\text{Non-linear term}},
    \label{eq:semilinear_eq}
\end{equation}
(derivation available in the Supplementary).
We analytically derive exact solutions for the linear term using the EI method and Taylor-expansions for the non-linear term that includes the $\x{0}$-predicting DBM $\mathbi{D}_{\boldsymbol{\theta}}(\x{t})$.

\subsection{Deriving the Solution to the Bridge SDE}
\label{subsec:sde_soln}
By simplifying and re-structuring the Bridge SDE (\Cref{eq:ddbm_backward_sde}, we observe that it also has a semi-linear structure.
We derive its exact solution by leveraging the EI method, which is particularly powerful for semi-linear differential equations that have a structure like \Cref{eq:semilinear_eq}.
By taking Taylor-expansions, we derive an efficient 1$^{\text{st}}$-order solution, allowing for an accurate sampling procedure:

\begin{proposition}
    Given an initial value $\x{s}$ and time steps $0 \leq t < s \leq T$, 
    a first-order approximation of the solution to $\x{t}$ (derived via the Taylor-expansion of the EI solution) is:
    \begin{align}
        \x{t} = \frac{\text{SNR}_s}{\text{SNR}_t} \frac{\alp{t}}{\alp{s}} & \x{s}  + \alp{t} \left( 1 - \frac{\text{SNR}_s}{\text{SNR}_t} \right) \mathbi{D}_{\boldsymbol{\theta}}(\x{s}) \notag \\
        & + \sig{t}\sqrt{1 - \frac{\text{SNR}_s}{\text{SNR}_t}} \, \boldsymbol{z}_t, 
    \label{eq:ddbm_sde_soln}
    \end{align}
    where $\boldsymbol{z}_t \sim \mathcal{N}(\mathbf{0}, \mathbf{I})$, and SNR$_t$ := $\nicefrac{\alp{t}^2}{\sig{t}^2}$ is the signal-to-noise ratio at time $t$ (proof in Supplementary).
\label{prop:sde_soln}
\end{proposition}
\vspace{-1em}
Note that~\Cref{prop:sde_soln} is an efficient $1^\text{st}$-order Taylor-approximation (with error O($\Delta t^2$)) of the exact EI solution. While higher orders are possible, they provide marginal gains for reasons explained in~\Cref{subsec:devising_dbmsolver}. 

\subsection{Deriving the Solution to the Bridge PF ODE}
\label{subsec:ode_soln}
Having set grounds with \cref{prop:sde_soln}, we next focus on the Bridge PF ODE~(\Cref{eq:ddbm_backward_ode}).
Similar to the Bridge SDE analysis above, we show that the Bridge ODE also exhibits semi-linearity in its structure, which has been largely overlooked in prior work.
We take advantage of this property and derive a closed-form exact solution through the EI method.
We then utilize the change-of-variables method to reformulate the solution as an exponentially-weighted integral.
Finally, we analytically minimize discretization errors by approximating via the Taylor-expansion of this integral, yielding a fast and efficient sampling procedure, as presented in~\Cref{prop:ode_soln} with proof in Supplementary.

\begin{proposition}
    \textit{Given an initial value $\x{s}$ and time steps $0 \leq t < s < T$, the exact solution to $\x{t}$ is:}
    \begin{align}
    \x{t} = & \,\frac{\alp{t}}{\alp{s}} e^{2(\lambda_s - \lambda_t)} \sqrt{\frac{\rho(\lambda_t, \lambda_T)}{\rho(\lambda_s, \lambda_T)}} \, \x{s} \notag \\
    & + \frac{\alp{t}}{\alp{T}} e^{2(\lambda_T-\lambda_t)} \left[ 1-\sqrt{\frac{\rho(\lambda_t, \lambda_T)}{\rho(\lambda_s, \lambda_T)}} \right] \x{T} \notag \\
    & + \alp{t} \, e^{-2\lambda_t} \sqrt{\rho(\lambda_t, \lambda_T)} \underbrace{\int_{\lambda_s}^{\lambda_t}  \, \frac{e^{2\lambda} \, \mathbi{D}_{\boldsymbol{\theta}}(\x{\lambda})}{\sqrt{\rho(\lambda, \lambda_T)}} \, \dif{\lambda}}_\text{The Exponential Integral},
    \label{eq:exact_solution_ode}
    \end{align}
    where $\lam{t} := \log (\alp{t} / \sig{t})$ with the inverse function $t_\lambda(\cdot)$, and $\x{\lambda}:= \x{t_\lambda(\lambda)}$ is the change-of-variable form for $\lambda$, and $\rho(a, b) := e^{2(a - b)} - 1$.
    Intuitively, $\lambda_t$ can be thought of as half the log SNR at time $t$.
\label{prop:ode_soln}
\end{proposition}

We simplify the \textit{Exponential Integral} in~\Cref{eq:exact_solution_ode} by taking its $(k-1)^\text{th}$ Taylor-expansion:
\begin{align}
    & \int_{\lambda_a}^{\lambda_b} \, \frac{e^{2\lambda} \, \mathbi{D}_{\boldsymbol{\theta}}(\x{\lambda})}{\sqrt{\rho(\lambda, \lambda_T)}} \, \dif{\lambda} \approx \underbrace{\mathcal{O}((\lam{t}-\lam{s})^{k+1})}_\text{Error-Term (Omitted)} \,+ \notag \\
    & \,\,\, \sum_{n=0}^{k-1} \, \underbrace{\mathbi{D}_{\boldsymbol{\theta}}^{(n)}(\x{\lambda_s})}_\text{Estimated} \underbrace{\int_{\lambda_s}^{\lambda_t}   \frac{e^{2\lambda}}{\sqrt{\rho(\lambda, \lambda_T)}} \, \frac{(\lambda-\lambda_s)^n}{n!} \dif{\lambda}}_\text{Analytically Computed (Supplementary)}
\label{eq:exact_solution_taylor}
\end{align}
where $k \geq 1$, and $\mathbi{D}_{\boldsymbol{\theta}}^{(n)}(\x{\lambda_s}) := \frac{\text{d}^n{\mathbi{D}_{\boldsymbol{\theta}}(\x{\lambda_s})}}{\dif{\lambda^n}}$ is the $n^\text{th}$-order derivative of $\mathbi{D}_{\boldsymbol{\theta}}(\cdot)$ \wrt $\, \lambda$.
Note that we omit the error term $\mathcal{O}((\lam{t}-\lam{s})^{k+1})$.
Notably, the $k=1$ (1st-order) approximation of our solution is equivalent to DBIM-1~\cite{dbim}.

\subsection{Devising DBMSolver using Equations~\ref{eq:ddbm_sde_soln} and~\ref{eq:exact_solution_ode}}
\label{subsec:devising_dbmsolver}

\paragraph{Initial Step.}
We observe that our solution of the Bridge PF ODE in~\Cref{eq:exact_solution_ode} is only valid for time $t < T$, which implies that it cannot be employed for an initial step from $s = T$ to time $t < T$.
This is because, as $s \rightarrow T$, $\rho(\lam{s}, \lam{T}) \rightarrow \rho(\lam{T}, \lam{T}) = 0$ which would cause the coefficient of $\x{s}$ to diverge to infinity.
We instead employ the $1^\text{st}$-order Bridge SDE solution~(\Cref{eq:ddbm_sde_soln}) exclusively for this initial step from $s = T$ to $t = T - \epsilon$ ($\epsilon \approx$ 1e-4), and the regular Bridge PF ODE solution~(\Cref{eq:exact_solution_ode}) for the subsequent steps (where $t < s \leq T - \epsilon$), as described next.

\vspace{-0.5em}
\paragraph{Subsequent Steps.}
Higher-order formulations of~\Cref{eq:exact_solution_taylor} can lead to a sampling procedure capable of generating high-quality images more efficiently, as demonstrated in previous work~\cite{dpm, dpmpp, unipc}.
This improvement stems from the fact that higher-order Taylor-approximations have smaller error bounds, yielding more accurate approximations.
Following this idea, we set $k = 2$, resulting in a $2^\text{nd}$-order Bridge ODE solution for~\Cref{eq:exact_solution_ode}.
We adopt this $2^\text{nd}$-order formulation for DBMSolver (complete derivation in Supplementary) and provide the rationale below.

\vspace{-0.5em}
\paragraph{Summarizing DBMSolver's Algorithm.}
Given time steps $T = t_N > t_{N-1} > \dots > t_1 > t_0 = 0$, we first compute $\xtilde{t_{N-1}}$ from prior image $\x{t_N} \sim p_T(\x{})$ using~\Cref{eq:ddbm_sde_soln}. For the next $N - 2$ steps, we iteratively apply~\Cref{eq:exact_solution_ode} with $k = 2$, yielding better approximations for each intermediate noisy sample until $\xtilde{t_1}$. 
To obtain the final $\xtilde{0}$ prediction, we solve the Bridge PF ODE from $t_1$ to $t_0$ using the widely used Euler method, resulting in a high-fidelity output.
We summarize it in~\Cref{alg:dbmsolver} and validate it empirically in the next section.

\vspace{-0.5em}
\paragraph{Rationale for DBMSolver Order Selection.}
\label{para:rationale_for_dbmsolver}
As mentioned above, our method involves two distinct integration phases.
For the \textbf{initial Bridge SDE step}, we employ the $1^\text{st}$-order solution~\Cref{prop:sde_soln}.
Our rationale is that this step is taken over a very small interval $\text{d}t = \epsilon$, rendering the marginal accuracy gains from a higher-order approximation unnecessary. Nonetheless, for completeness, we present derivations for higher-order Bridge SDE solutions in the Supplementary.

For the \textbf{subsequent Bridge ODE steps}, we note that $k \geq 3$ involves a \textit{non-elementary antiderivative}, meaning its solution cannot be expressed in closed form using elementary functions (e.g., polynomials, exponentials, or logarithms).
While such equations can be addressed using numerical techniques like linear multistep methods, an approach adopted by DBIM~\cite{dbim} for their higher-order (DBIM-2 and DBIM-3) samplers, we instead avoid this associated complexity and larger error bounds. 
As the experiments in \Cref{sec:exp_results} will show, our 2nd-order DBMSolver significantly outperforms~\cite{dbim}'s 1st-order DBIM-1 and the numerically approximated higher-order DBIM-2 and DBIM-3.
Therefore, we restrict DBMSolver to the $k=2$ ($2^\text{nd}$-order) solution, which remains analytically tractable. Ablations in Supplementary confirm $k=2$ generates better FID scores than $k=1$ (DBIM-1's analog).

\begin{algorithm}[t!]
\caption{DBMSolver: A Training-free Sampler for Diffusion-based I2I Translation}
\begin{algorithmic}
\STATE \textbf{Inputs:} Pretrained DBM $\mathbi{D}_{\boldsymbol{\theta}}(\cdot)$, Number of sampling steps $N$, Time steps $T=t_N>\cdots>t_1>t_0=0$, and Prior distribution $p_T(\mathbf{x})$.
\vspace{0.5em}
\STATE \textbf{Initialization:} Sample $\xtilde{T} \sim p_T(\mathbf{x})$, $\boldsymbol{z} \sim \mathcal{N}(\mathbf{0}, \mathbi{I})$, and $\xtilde{0} \leftarrow \mathbi{D}_{\boldsymbol{\theta}}(\xtilde{t_N})$
\vspace{0.5em}
\STATE \textbf{Initial Stochastic Update:} Compute $\xtilde{t_{N-1}}$ from $\xtilde{T}$ using~\cref{eq:ddbm_sde_soln}.
\vspace{0.5em}
\STATE \textbf{Subsequent Deterministic Refinement:}
\FOR{$i=N-1$ \textbf{to} $1$}
\IF{$i > 1$}
\STATE $a \leftarrow t_i$, and $b \leftarrow t_{i-1}$.
\STATE Compute $\xtilde{b}$ from $\xtilde{a}$ using~\cref{eq:exact_solution_ode} with $k = 2$.
\ELSE
\STATE $\mathbf{s}_{\boldsymbol{\theta}} \leftarrow \text{GetScoreFromX0}(\mathbi{D}_{\boldsymbol{\theta}}(\xtilde{t_1}), \xtilde{t_1}, t_1, \x{T})$ \hfill \COMMENT{{$\triangleright$ Convert $\x{0}$-pred to score via \cref{eq:score_to_x0}}}
\STATE $\mathbf{s}_{\text{trans}} \leftarrow \score{p_{t_1}}{\x{T} \mid \xtilde{t_1}}$ \hfill \COMMENT{{$\triangleright$ Tractable transition score from \cref{eq:score_of_h_function}}}
\STATE $\dif{\x{t_1}} \leftarrow \driftConst{\xtilde{t_1}, t_1} - \frac{1}{2} \diffConst{t_1}^2 \mathbf{s}_{\boldsymbol{\theta}} + \diffConst{t_1}^2 \mathbf{s}_{\text{trans}}$
\STATE $\xtilde{0} \leftarrow \xtilde{t_1} + (t_0 - t_1) \, \dif{\x{t_1}}$
\hfill \COMMENT{{$\triangleright$ Final Euler Update}}
\ENDIF
\ENDFOR
\STATE \textbf{Output: $\xtilde{0}$}
\hfill \COMMENT{{$\triangleright$ Final Translated Image}}
\end{algorithmic}
\label{alg:dbmsolver}
\end{algorithm}

\begin{figure*}[t!]
\vspace{-1em}
    \centering
    \includegraphics[width=0.8\linewidth]{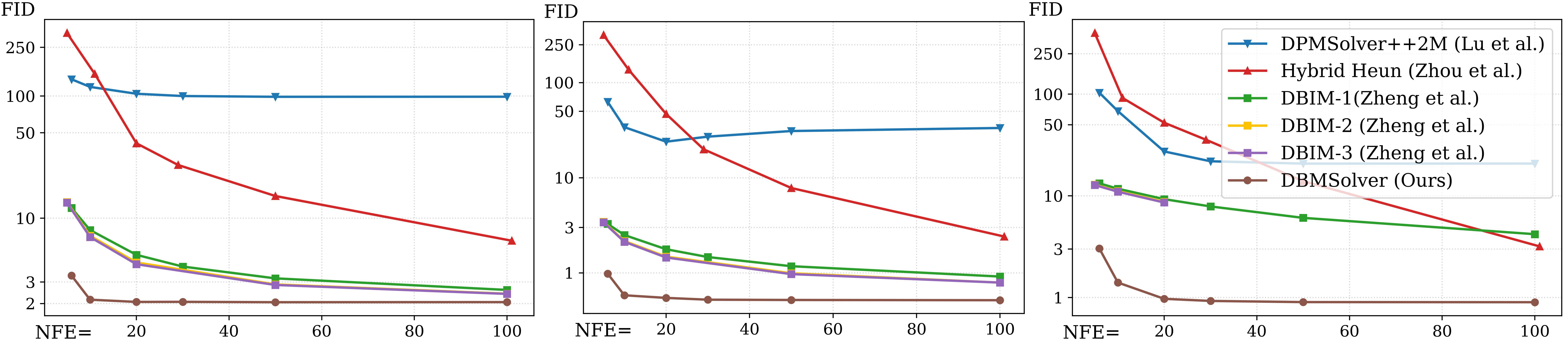}
    \begin{minipage}{0.99\linewidth}
        \centering
        \subcaptionbox{DIODE \imgres{256}}[0.32\linewidth]{}
        \hspace{-4.5em}
        \subcaptionbox{Edges2Handbags \imgres{64}}[0.32\linewidth]{}
        \hspace{-3.5em}
        \subcaptionbox{Face2Comics \imgres{256}}[0.32\linewidth]{}
    \end{minipage}
    \caption{FID vs. NFE on DIODE, E2H, and Face2Comics datasets. We consistently get lower FID scores with fewer NFEs.}
    \label{fig:trends}
    \vspace{-1em}
\end{figure*}

\section{Experiments and Results}
\label{sec:exp}
We conducted extensive experiments to evaluate DBMSolver against established baselines on various I2I Translation tasks, including conditional image inpainting and semantics-to-image generation, to demonstrate its versatility across diverse tasks.
Specifically, we evaluate the following datasets: Sketch-to-Image on \textit{Edges2Handbags} (E2H)~\cite{pix2pix}, Surface normals-to-Image on \textit{DIODE-Outdoor}~\cite{diode}, Face-to-Comic stylization on \textit{Face2Comics} (F2C), Conditional Image Inpainting with central masks on \textit{ImageNet}~\cite{imagenet}, and Semantic Label-to-Face generation on \textit{CelebAMask-HQ}~\cite{celebamaskhq}.
We benchmark against DDBM (Hybrid Heun)~\cite{ddbm}, DBIM variants (1st/2nd/3rd-order)~\cite{dbim}.
Note that, DBIM-2/3 use their multistep numerics for higher order.
Additional baselines of DDIB~\cite{ddib}, SDEdit~\cite{sdedit}, and I$^\text{2}$SB~\cite{i2sb} are evaluated following the DDBM and DBIM protocols.

We mainly assess sampling quality using FID~\cite{fid}, MSE, IS, and LPIPS, and computational efficiency via the number of forward evaluations NFEs~\cite{ddim,dpm}, following prior works~\cite{ddbm, dbim}. 
For CelebAMask-HQ, we additionally report the classification accuracy (CA), following~\cite{bbdm}.
We use the publicly available DBM checkpoints from~\cite{ddbm} for E2H and DIODE, highlighting DBMSolver's training-free integration. 
For ImageNet inpainting, we adopt the DBM checkpoint from~\cite{dbim}, which was finetuned via I$^\text{2}$SB~\cite{i2sb} from a pre-trained N2I Diffusion Model. 
For datasets lacking checkpoints (e.g., Face2Comics, CelebAMask-HQ), we train DBMs from scratch using the ADM U-Net~\cite{dmdhariwal}, following standard diffusion architectures. 
We describe the training and sampling details in the Supplementary.

\begin{table}[b!]
    \centering
    \caption{
            Quantitative results on DIODE \imgres{256}~\cite{diode}. FID ($\downarrow$), IS ($\uparrow$), MSE($\downarrow$), and LPIPS ($\downarrow$) are reported against NFE ($\downarrow$). 
            }
    \resizebox{0.9\columnwidth}{!}{
    \begin{tabular}{@{}clcc*4{c}@{}}
        \toprule
         \multirow{2}{*}{Family} & \multirow{2}{*}{Method} & \multirow{2}{*}{NFE $\downarrow$} & & \multicolumn{4}{c}{DIODE \imgres{256}~\cite{diode}} \\
         \cmidrule{5-8}
         &&&&FID$\downarrow$&IS $\uparrow$&LPIPS$\downarrow$&MSE$\downarrow$\\
         \midrule
         \multirow{3}{0.15\linewidth}{\centering Diffusion \& Flow}&DDIB~\cite{ddib}& $\geq$40 & &242.3 &4.22 &0.798 &0.794\\
         &SDEdit~\cite{sdedit}& $\geq$40 & &31.14 &5.70& 0.714 &0.534\\
         &I$^\text{2}$SB~\cite{i2sb}& $\geq$40 & & 9.34& 5.77 &0.373&0.145\\
         \midrule
         \multirow{7}{0.15\linewidth}{\centering DBM Sampling}&Hybrid Heun~\cite{ddbm}& 119 & &4.43&\textbf{6.21}&0.244&0.084 \\
         & DBIM-1~\cite{dbim} & 20 & &4.99 &6.10& 0.201 &0.017\\
         & DBIM-2~\cite{dbim} & 20 & &4.40&6.11&0.200&0.017 \\
         & DBIM-3~\cite{dbim} & 20 & &4.23&6.05&0.201&0.017\\
         & ODES3~\cite{odes3} & 28 & &2.29&5.92&0.203&0.018\\
         \cmidrule{2-8}
         \rowcolor{gray!20} 
         \cellcolor{white!0}& DBMSolver \textbf{(Ours)} & 6 & &3.38&6.00&\textbf{0.196}& \textbf{0.015}\\
         \rowcolor{gray!20} 
         \cellcolor{white!0}& DBMSolver \textbf{(Ours)} & 20 & &\textbf{2.06}&6.00&0.198&0.018 \\
         \bottomrule
    \end{tabular}
    }
    \label{tab:diode_comparison}
\end{table}

\begin{table}[t!]
    \centering
    \caption{
            Quantitative results on E2H~\imgres{64}~\cite{pix2pix}. FID ($\downarrow$), IS ($\uparrow$), MSE($\downarrow$), and LPIPS ($\downarrow$) are reported against NFE ($\downarrow$). 
            }
    \resizebox{\columnwidth}{!}{
    \begin{tabular}{@{}clcc*4{c}@{}}
        \toprule
         \multirow{2}{*}{Family} & \multirow{2}{*}{Method} & \multirow{2}{*}{NFE $\downarrow$} & & \multicolumn{4}{c}{Edges2Handbags\imgres{64}~\cite{pix2pix}} \\
         \cmidrule{5-8}
         &&&&FID$\downarrow$&IS $\uparrow$&LPIPS$\downarrow$&MSE$\downarrow$\\
         \midrule
         \multirow{3}{0.15\linewidth}{\centering Diffusion \& Flow}&DDIB~\cite{ddib}& $\geq$40 & &186.84 &2.04 &0.869& 1.050 \\
         &SDEdit~\cite{sdedit}& $\geq$40 & &26.50 &3.58 &0.271 &0.510\\
         &I$^\text{2}$SB~\cite{i2sb}& $\geq$40 & &7.43 &3.40 &0.244 &0.191\\
         \midrule
         \multirow{7}{0.15\linewidth}{\centering DBM Sampling}&Hybrid Heun~\cite{ddbm}& 119 & & 1.83 &\textbf{3.73}& 0.142 &0.040 \\
         & DBIM-1~\cite{dbim} &  20& & 1.74 &3.63 &\textbf{0.095} &\textbf{0.005} \\
         & DBIM-2~\cite{dbim} &  20& &1.48&3.60&0.098& \textbf{0.005} \\
         & DBIM-3~\cite{dbim} &  20& &1.45&3.61&0.098& \textbf{0.005} \\
         & ODES3~\cite{odes3} &  28& & 0.54 & 3.65 & 0.097 & \textbf{0.005} \\
         \cmidrule{2-8}
         \rowcolor{gray!20} 
         \cellcolor{white!0} & DBMSolver \textbf{(Ours)} &  6& &0.93&3.60&0.106& 0.006\\
         \rowcolor{gray!20} 
         \cellcolor{white!0} & DBMSolver \textbf{(Ours)} & 20 &&\textbf{0.53}&3.64&0.099&\textbf{0.005}\\
         
         \bottomrule
    \end{tabular}
    }
    \label{tab:e2h_comparison}
\end{table}

\begin{table}[t!]
\captionof{table}{\textbf{Left:} Quantitative comparison on Face2Comics~\cite{face2comics}. \textbf{Right:}  Quantitative results for Label-to-Face Generation on CelebAMask-HQ~\cite{celebamaskhq} at NFEs of 6 and 30, complementing the visual examples in \Cref{fig:celeba_qualitative}.}
    \centering
    \begin{minipage}[t!]{0.53\linewidth}
        \vspace{-0.5em}
        \resizebox{!}{0.6\linewidth}{
        \begin{tabular}{@{}lcc@{}}
        \toprule
        \multicolumn{3}{c}{Face2Comics \imgres{256}~\cite{face2comics}} \\
        \midrule
        Method & NFE $\downarrow$ & FID $\downarrow$ \\
        \midrule
        \rowcolor{gray!10} 
        \multicolumn{3}{l}{\textit{GANs \& Other Diffusion-based Models:}} \\
        \midrule
        Pix2Pix~\cite{pix2pix} & 1 & 49.96 \\
        CycleGAN~\cite{cyclegan} & 1 & 35.13 \\
        DRIT++~\cite{dritpp} & -- & 28.87 \\
        CDE~\cite{cde} & -- & 33.98 \\
        LDM~\cite{ldm} & -- & 24.28 \\
        BBDM~\cite{bbdm} & 200 & 23.20 \\
        \midrule
        \rowcolor{gray!10} 
        \multicolumn{3}{l}{\textit{DBM Sampling:}} \\
        \midrule
        Hybrid Heun~\cite{ddbm} & 119 & 2.36 \\
        DBIM-1~\cite{dbim} & 20 & 9.28 \\
        DBIM-2~\cite{dbim} & 20 & 8.74 \\
        DBIM-3~\cite{dbim} & 20 & 8.61 \\
        \midrule
        \rowcolor{gray!20} 
        DBMSolver \textbf{(Ours)} & 6 & 3.04 \\
        \rowcolor{gray!20} 
        DBMSolver \textbf{(Ours)} & 20 & \textbf{0.96} \\
        \bottomrule
        \end{tabular}
        }
        \label{tab:f2c_comp}
    \end{minipage}
    \hspace{-1em}
    \centering
    \begin{minipage}{0.46\linewidth}
        \vspace{-0.5em}
        \resizebox{\linewidth}{!}{
        \begin{tabular}{@{}lcc@{}}
        \toprule
        \multicolumn{3}{c}{CelebAMask-HQ \imgres{256}~\cite{celebamaskhq}} \\
        \midrule
        Method & NFE $\downarrow$ & FID $\downarrow$\\ 
        \midrule
        \rowcolor{gray!10} 
        \multicolumn{3}{l}{\textit{GANs \& Other Diffusion-based Models:}}\\
        \midrule
        Pix2Pix~\cite{pix2pix} & 1 & 56.99 \\
        CycleGAN~\cite{cyclegan} & 1 & 78.23 \\
        DRIT++~\cite{dritpp} & -- & 77.79 \\
        SPADE~\cite{spade} & -- & 44.17 \\
        OASIS~\cite{oasis} & -- & 27.75 \\
        CDE~\cite{cde} & -- & 24.40 \\
        LDM~\cite{ldm} & -- & 22.81 \\
        BBDM~\cite{bbdm} & 200 & 21.35 \\
        \midrule
        \rowcolor{gray!10} 
        \multicolumn{3}{l}{\textit{DBM Sampling:}} \\
        \midrule
        Hybrid Heun~\cite{ddbm} & 119 & 97.75 \\
        DBIM-1~\cite{dbim}& 20 & 23.41 \\
        DBIM-2~\cite{dbim}& 20 & 19.86 \\
        DBIM-3~\cite{dbim}& 20 & 19.49 \\
        \midrule
        \rowcolor{gray!20} 
        DBMSolver \textbf{(Ours)}& 20 & \textbf{17.56} \\        
        \bottomrule
    \end{tabular}
        }
        \label{tab:celeba_comparison}
    \end{minipage}
    \vspace{-0.5 em}
\end{table}
\begin{table}[t!]
    \begin{minipage}{\linewidth}
        \centering
        \captionof{table}{Quantitative results for Class-Conditional Inpainting (center $128 {\times} 128$ mask) on ImageNet~\cite{imagenet}. DBMSolver achieves superior FID and Classification Accuracy (CA) across all NFEs, delivering high image fidelity with only 6 NFEs, outperforming prior methods that require more NFEs for comparable quality.}
        \resizebox{0.8\linewidth}{!}{
        \begin{tabular}{@{}lccccc@{}}
            \toprule
            \multirow{2}{*}{Methods} &  \multicolumn{5}{c}{ImageNet\imgres{256}~\cite{imagenet}} \\
            \cmidrule{2-6}
            & Time $\downarrow$ & Rate $\uparrow$  & NFE $\downarrow$ & FID $\downarrow$ & CA $\uparrow$\\ 
            \midrule
            \rowcolor{gray!10} 
            \multicolumn{6}{l}{\textit{Other Diffusion-based Models:}}\\
            \midrule
            DDRM~\cite{ddrm} & -- & -- & 20 & 24.40 & 62.1 \\
            $\Pi$GDM~\cite{pigdm} & -- & -- & 100 & 7.30 & 72.6 \\
            DDNM~\cite{ddnm} & -- & -- & 100 & 15.10 & 55.9 \\
            Palette~\cite{palette} & -- & -- & 1000 & 6.10 & 63.0 \\
            I$^\text{2}$SB~\cite{i2sb} & -- & -- & 1000 & 4.90 & 66.1 \\
            \midrule
            \rowcolor{gray!10} 
            \multicolumn{6}{l}{\textit{Sampling via Diffusion Bridge Models:}}\\
            \midrule
            Hybrid Heun~\cite{ddbm} & 172.78 & 0.95 & 119 & 6.02 & 69.5 \\
            DBIM-1~\cite{dbim} & 13.67 & 12.19 & 20 & 4.13 & 71.9 \\
            DBIM-2~\cite{dbim} & 13.67 & 12.20 & 20 & \textbf{4.07} & \textbf{72.0} \\
            DBIM-3~\cite{dbim} & 13.61 & 12.24 & 20 & \textbf{4.07} & \textbf{72.0} \\
            \midrule
            \rowcolor{gray!20} 
            DBMSolver \textbf{(Ours)} & 3.66 & 45.41 & 6 & 4.98 & 70.8 \\
            \rowcolor{gray!20} 
            DBMSolver \textbf{(Ours)} & 14.05 & 11.85 & 20 & \textbf{4.07} & \textbf{72.0} \\
            \bottomrule
        \end{tabular}
        }
        \label{tab:inpainting_comparison}
            \vspace{-1em}
    \end{minipage}
\end{table}

\subsection{Results}
\label{sec:exp_results}

\paragraph{Image Translation on E2H $\mathbf{(64 {\times} 64)}$ and DIODE $\mathbf{(256 {\times} 256)}$.}
\Cref{tab:diode_comparison} reports FID scores and NFEs across methods. 
DBMSolver achieves state-of-the-art results with significantly fewer evaluations. 
At just 6 NFEs, it achieves FID scores of 0.97 (E2H) and 3.38 (DIODE), outperforming Hybrid Heun and DBIM-1/2/3. 
Its high efficiency at low NFEs enables rapid sampling, making it well-suited for real-time DBM applications by supporting faster generation and higher throughput.
It exhibits strong scalability with increasing NFEs, yielding further improvements in FID.

\begin{figure*}[t!]
    \begin{minipage}{\textwidth}
        \centering
            \centering
            \setcounter{subfigure}{0} 
            \includegraphics[width=0.9\linewidth]{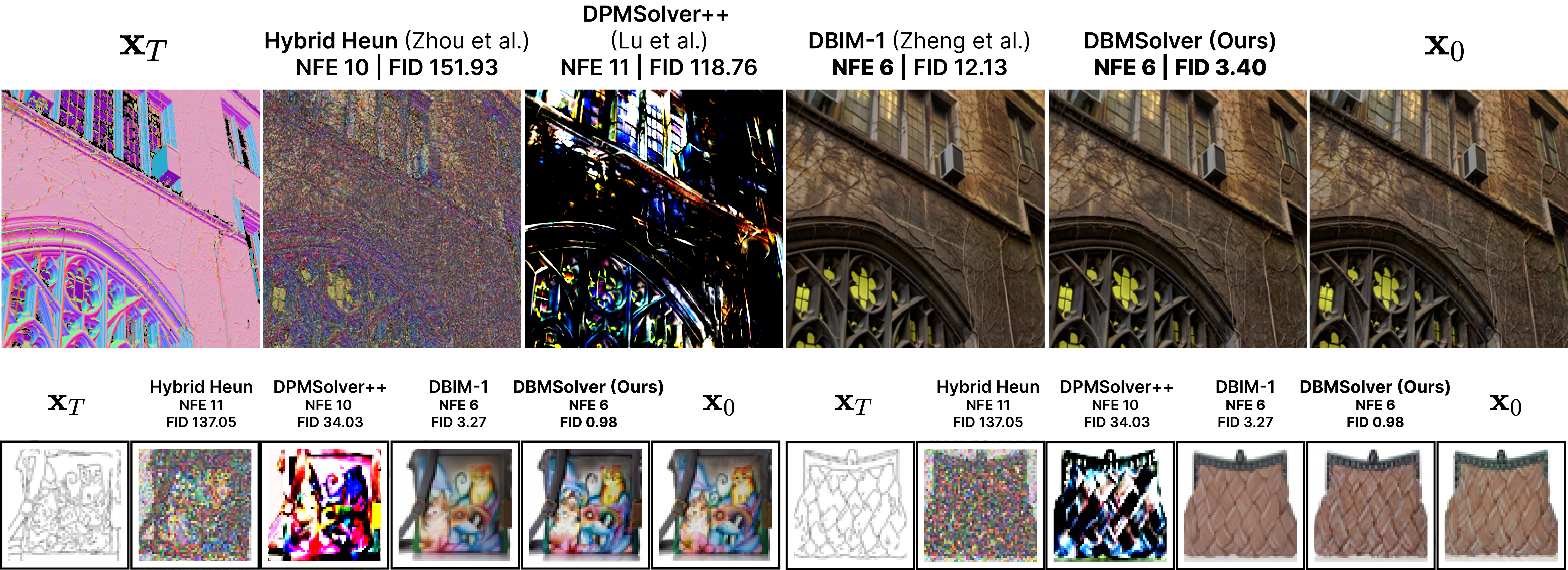}
            \label{fig:combined_qual}
        \vspace{-0.5em}
        \captionof{figure}{
        Visuals for~\Cref{tab:diode_comparison,tab:e2h_comparison} (DPMSolver++ and HH shown at 11 NFEs due to poor 6-NFE quality).}
        \label{fig:dbmsolver_comparison}
    \end{minipage}
\vspace{-1em}
\end{figure*}

The trends in \Cref{fig:trends}-a,b show that as NFE increases, DBMSolver quickly achieves high fidelity and remains stable. \Cref{fig:dbmsolver_comparison} supports this, indicating that even at low NFEs (e.g., 6), DBMSolver and DBIM generate visually rich, coherent outputs, outperforming others in detail and realism. DPMSolver++2M preserves structure but lacks vibrant colors and texture, especially at lower NFEs. 
While DBIM yields appealing outputs, it lacks fine detail compared to our method-- a gap reflected in FID and trend metrics. Please refer to the intricate structural details observable in the tree branches and twigs within the DIODE images, as well as the fine-grained textures and contours present in the handbag depictions. DBMSolver consistently balances efficiency and quality across all datasets.

\begin{figure*}[t!]
    \centering
    \includegraphics[width=0.9\linewidth]{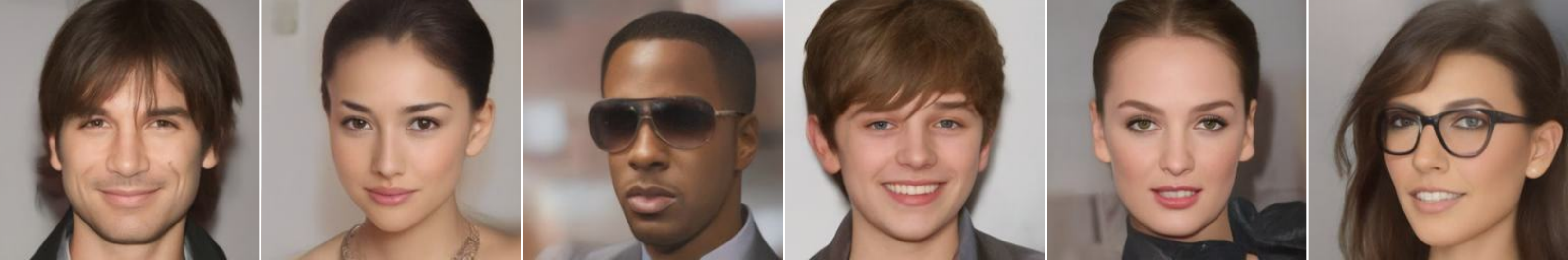}
    \vspace{-0.5em}
    \captionof{figure}{Generated samples on CelebAMask-HQ \imgres{256} using our DBMSolver in 6 NFEs.}
    \label{fig:oldteaser}
    \vspace{-1em}
\end{figure*}
\begin{figure*}[t!]
    \centering
    \includegraphics[width=0.9\linewidth]{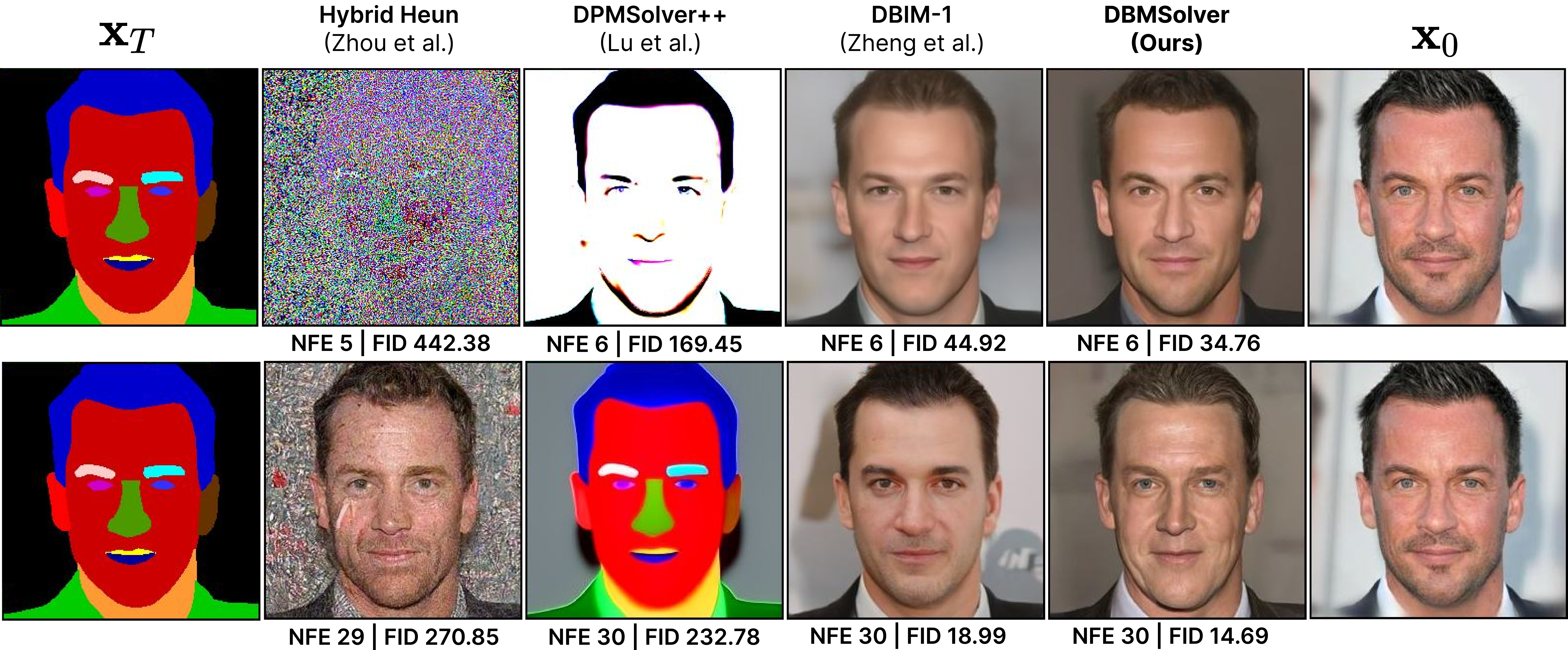}
    \vspace{-0.5em}
    \caption{Label-to-Face Generation on CelebAMask-HQ \imgres{256}.}
    \label{fig:celeba_qualitative}
    \vspace{-1.5em}
\end{figure*}

\vspace{-1em}
\paragraph{Label-to-Face on CelebAMask-HQ $\mathbf{(256 {\times} 256)}$.}
\Cref{fig:celeba_qualitative,fig:oldteaser} together with \Cref{tab:celeba_comparison} show that our method generates images with precise facial segmentation and coherent boundaries. 
At as low as 6 NFEs, DBMSolver achieves an FID of 34.76 (in \Cref{fig:celeba_qualitative}) outperforming DBIM-1's 44.92 as well as GAN-based models and other diffusion approaches, while using significantly fewer NFEs.
Visually, DBMSolver preserves fine structural details such as eye contours, hairlines, and mask edges, which are often blurred or distorted in DBIM outputs.
DBMSolver consistently produces sharper, more anatomically faithful generations, enhancing both realism and image accuracy.

\begin{figure*}[t!]
\centering
    \includegraphics[width=0.9\linewidth]
    {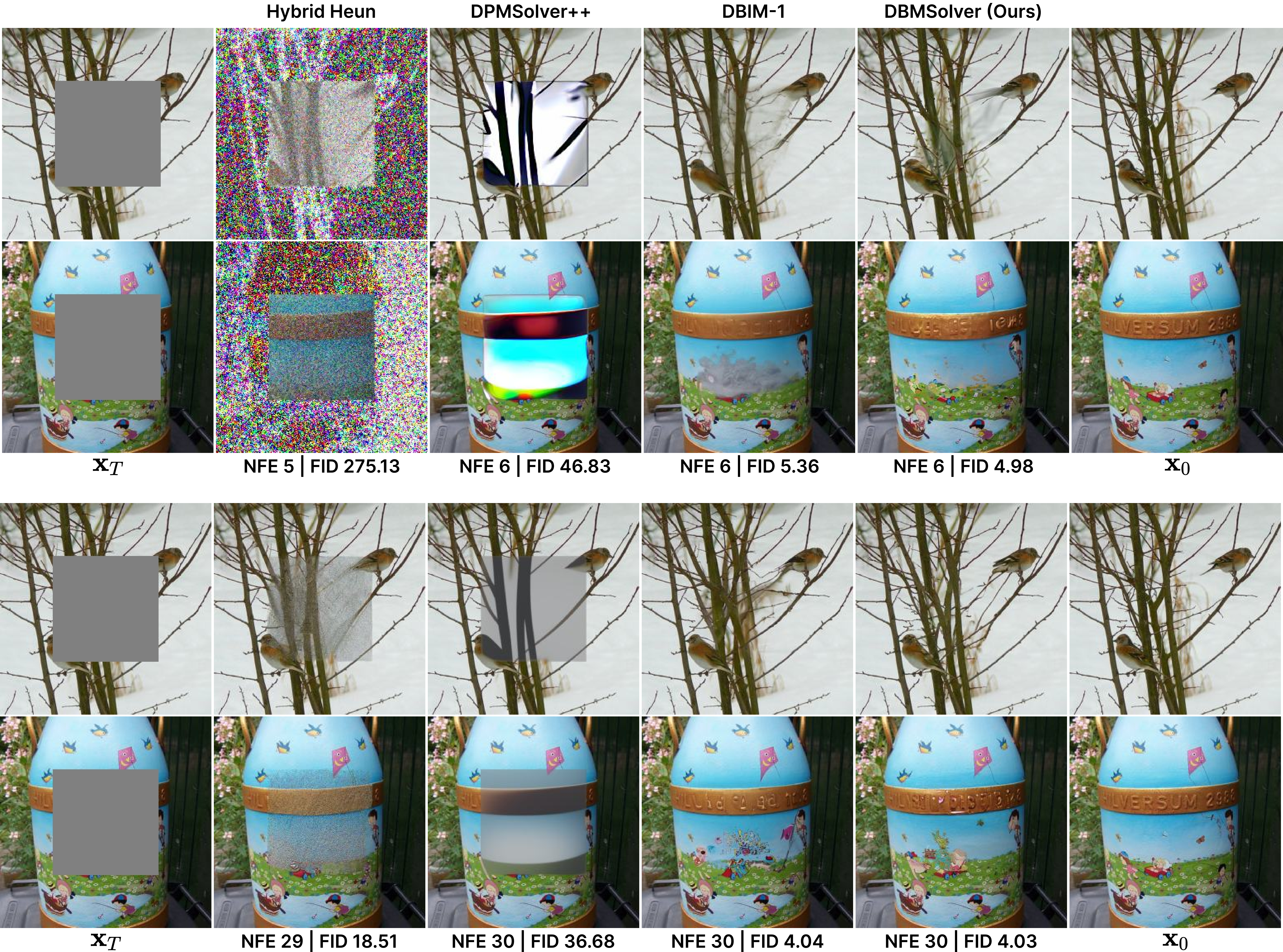}
    \vspace{-0.5em}
    \caption{Class-Conditional Inpainting on Images of ImageNet dataset \imgres{256}.}
    \label{fig:imagenet_qualitative}
    \vspace{-1em}
\end{figure*}

\begin{figure}[t!]
    \centering
    \begin{minipage}[t!]{\linewidth}
        \centering
        \includegraphics[width=\linewidth]{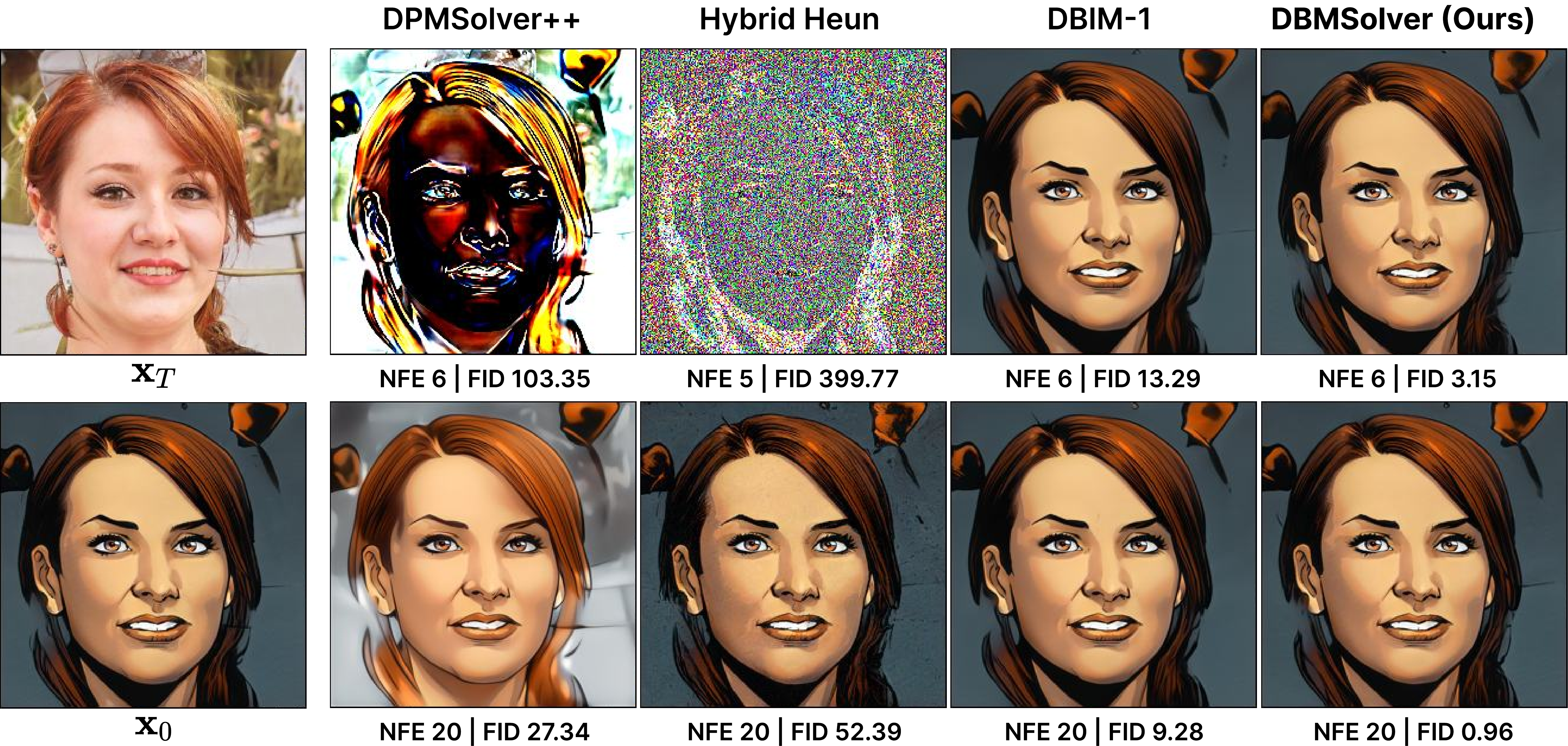}
        \vspace{-1.5em}
        \captionof{figure}{Image Stylization on Face2Comics \imgres{256}.}
        \vspace{-1.5em}
    \label{fig:qualitative_comparison}
    \end{minipage}
\end{figure}

\vspace{-1em}
\paragraph{Image Stylization on Face2Comics $\mathbf{(256 {\times} 256)}$.}
As shown in Tables~\cref{tab:f2c_comp} and~\ref{fig:qualitative_comparison}, DBMSolver achieves top performance with just 10 NFEs. At 20 NFEs, it attains an FID of 0.96, outperforming HH (2.36 at 119 NFEs), DBIM-2 (8.74), and various GAN and diffusion methods. This highlights DBMSolver's efficiency and sample quality across diverse datasets, as illustrated in~\Cref{fig:teaser}. Even at 6 NFEs, its outputs rival those of higher-NFE baselines, demonstrating strong perceptual fidelity at minimal cost.

\vspace{-1.5em}
\paragraph{Class-Conditional Inpainting on ImageNet $\mathbf{(256 {\times} 256)}$.}
\Cref{tab:inpainting_comparison} and~\Cref{fig:imagenet_qualitative} demonstrate DBMSolver's superior performance. At just 6 NFEs, it delivers coherent structure and textures with FID 4.98), outperforming DBIM-1's 5.36 FID.  
DBMSolver maintains fidelity and avoids the blurry textures seen in DBIM-1, evident in the milk barrel writings and hallucinated drawings.
DBIM and HH results were reproduced using their official code.

\vspace{-1em}
\paragraph{Relation to Schrödinger Bridges.} 
We distinguish our work from methods addressing the Schrödinger Bridge (SB) problem, such as~\cite{btts}. 
SB methods solve a specific, entropy-regularized optimal transport problem; \cite{btts}, for example, assumes a ``tractable" SB formulation that results in a specific set of bridge SDEs/ODEs.
Our approach is distinct. 
DBMSolver is a general-purpose solver derived specifically for the generalized VP/VE-Bridge framework of~\cite{ddbm}, which unifies VP, VE, and Brownian bridge constructions. 
Because DBMs'~\cite{ddbm} underlying framework is different from SB, our resulting Bridge SDEs/ODE solutions (Propositions~\ref{prop:sde_soln} and~\ref{prop:ode_soln}) are also fundamentally different. 
Therefore, while~\cite{btts} derives solutions for their specific SB equations, DBMSolver is a novel solver tailored to the SDEs/ODEs of the generalized VP/VE-Bridge~\cite{ddbm} framework.

\begin{table}[t!]
\centering
\caption{DBMSolver (training-free) vs. distillation (retrained) on equiv. quality (FID at low NFE). CDBM~\cite{cdbm}/IBMD~\cite{invbridge} are close at 1-2 NFEs, but \textbf{require considerable training overhead}.}
\label{tab:distill_comp}
\vspace{-0.5em}
\resizebox{\linewidth}{!}{
\begin{tabular}{lcccc}
\toprule
Task and Resolution & Method & NFE $\downarrow$ & FID $\downarrow$ & Training-free? \\
\midrule
E2H  & CDBM~\cite{cdbm} & 2 & 1.30 & {\color{Red}\ding{55}} \\
\imgres{64} & IBMD~\cite{invbridge} & 1 & 1.26 & {\color{Red}\ding{55}} \\
\rowcolor{gray!20} & DBMSolver \textbf{(Ours)} & 6 & \textbf{0.97} & {\color{ForestGreen}\ding{51}} \\
\midrule
DIODE & CDBM~\cite{cdbm} & 2 & 3.66 & {\color{Red}\ding{55}} \\
\imgres{256} & IBDM~\cite{invbridge} & 1 & 4.07& {\color{Red}\ding{55}}\\
\rowcolor{gray!20} & DBMSolver \textbf{(Ours)} & 6 & \textbf{3.38} & {\color{ForestGreen}\ding{51}} \\
\midrule
ImageNet Inpainting & CDBM~\cite{cdbm} & 2 & 5.65 & {\color{Red}\ding{55}} \\
\imgres{128} & IBMD~\cite{invbridge} & 1 & 5.87 & {\color{Red}\ding{55}} \\
\rowcolor{gray!20} & DBMSolver \textbf{(Ours)} & 6 & \textbf{4.98}* & {\color{ForestGreen}\ding{51}} \\
\bottomrule
\end{tabular}
}
\vspace{-1em}
\end{table}

\section{Conclusion}
\label{sec:conclusion}
We introduce DBMSolver, a principled, training-free sampler that significantly enhances the efficiency and quality of diffusion bridge-based I2I translation. By leveraging the semi-linear structure of the Bridge SDE and PF ODE, DBMSolver accelerates sampling without compromising fidelity. Experiments on diverse datasets such as Edges2Handbags, DIODE-Outdoor, Face2Comics, CelebAMask-HQ, and ImageNet Inpainting show that DBMSolver 
sets a new benchmark for efficient diffusion bridge models and marking a step toward the practical deployment of powerful solvers for I2I Translation.

\vspace{-1em}
\paragraph{Limitations and Future Work.}
A limitation is that DBMSolver performed similarly to previous solvers on more realistic tasks like ImageNet Inpainting, which we hypothesize is due to the non-linear $\boldsymbol{D_\theta}$ term being the main cause of approximation errors.
Exploring DBMs for text-conditioned I2I translation, adaptive stepsize for DBM sampling, or integration with flow-matching~\cite{flowmatch}, are also promising avenues for research.

\section*{Acknowledgements}
{\small This work was partly supported by the InnoCORE program (26-InnoCORE-01), the IITP grants (RS-2022-II220077, RS-2022-II220113, RS-2022-II220959, RS-2022-II220871, RS-2021-II211343 (SNU AI), RS-2025-25442338 (AI Star Fellowship-SNU)) funded by the Korea government (MSIT), grants (RS-2025-25462891 (US-KOR BARI), RS-2025-25453780) funded by MOTIR, a grant of Korean ARPA-H Project through the Korea Health Industry Development Institute (KHIDI), funded by the Ministry of Health \& Welfare, Republic of Korea (RS-2025-25424639), and the BK21 FOUR program, SNU in 2025.}

{
    \small
    \bibliographystyle{ieeenat_fullname}
    \bibliography{main}
}

\clearpage


\section*{Supplementary material}

# DBMSolver: A Training-free Diffusion Bridge Sampler for High-Quality Image-to-Image Translation

Supplementary Material

## 1. Diffusion Models

The reverse diffusion process is given by the PF ODE [1, 11]:

$$\mathrm{d}\boldsymbol{x}_t = \left[ f(\boldsymbol{x}_t, t) - \frac{1}{2} g(t)^2 \nabla_{\boldsymbol{x}_t} \log p_t(\boldsymbol{x}) \right] \mathrm{d}t, \tag{1}$$

where the marginal distribution of $\boldsymbol{x}_t$ at $t$ is $p_t(\mathbf{x})$, and $\nabla_{\boldsymbol{x}_t} \log p_t(\mathbf{x})$ is its *score function*, which is learned by a neural network [4]. Furthermore, the drift and diffusion coefficients are:

$$f(\boldsymbol{x}_t, t) = \boldsymbol{x}_t \frac{\mathrm{d}}{\mathrm{d}t} \log \alpha_t, \text{ and } g(t)^2 = -2\sigma_t^2 \frac{\mathrm{d}}{\mathrm{d}t} \log \left( \frac{\alpha_t}{\sigma_t} \right),$$

where $\alpha_t := \alpha(t)$ and $\sigma_t := \sigma(t)$, for time $t \in [0, T]$ (where $T > 0$). Different formulations of $\alpha_t$ and $\sigma_t$ give rise to different formulations for the diffusion process. Prior works hand-design these to obtain the *variance-preserving* (VP) [11, 15], *variance-exploding* (VE) [6], and *TrigFlow* [9] formulations. We contrast the design choices of such diffusion formulations in Table 1.

Table 1. Design choices for widely-used diffusion formulations. DDBM [15] mainly utilizes the VP and VE bridges, while TrigFlow is utilized by [9].

| Formulation | $\alpha_t$ | $\sigma_t$ | $f(\boldsymbol{x}_t, t)$ | $g(t)^2$ | $\text{SNR}_t = \alpha_t^2/\sigma_t^2$ | Domain of $t$ |
|---|---|---|---|---|---|---|
| VP [11, 15] | $e^{-(0.5t^2+0.05t)}$ | $\sqrt{1-e^{-(t^2+0.1t)}}$ | $-(t+0.05)\boldsymbol{x}_t$ | $2t+0.1$ | $1/(e^{(t^2+0.1t)}-1)$ | $[0.0001, 1]$ |
| VE [6] | 1 | $t$ | $\mathbf{0}$ | $2t$ | $1/t^2$ | $[0.002, 80]$ |
| TrigFlow [9] | $\cos(t)$ | $\sin(t)$ | $-\tan(t)\,\boldsymbol{x}_t$ | $2\tan(t)$ | $\cot^2(t)$ | $[0, \pi/2]$ |

## 2. Proofs & Derivations

### 2.1. Proof of Proposition 1 of the main paper

Given a well-trained DBM $\boldsymbol{D}_\theta(\cdot)$ that approximates data sample $\boldsymbol{x}_0$, we can simplify Equation 2 of the main paper as:

$$
\begin{aligned}
\mathrm{d}\boldsymbol{x}_t &= \big( f(\boldsymbol{x}_t, t) - g(t)^2[\nabla_{\boldsymbol{x}_t}\log p_t(\boldsymbol{x}_t \mid \boldsymbol{x}_T) - \nabla_{\boldsymbol{x}_t}\log p_t(\boldsymbol{x}_T \mid \boldsymbol{x}_t)]\big)\,\mathrm{d}t + g(t)\,\mathrm{d}\mathbf{w}_t \\
&= \left( \boldsymbol{x}_t \frac{\mathrm{d}\log\alpha_t}{\mathrm{d}t} + 2\sigma_t^2 \frac{\mathrm{d}\log\left(\frac{\alpha_t}{\sigma_t}\right)}{\mathrm{d}t} \left[\nabla_{\boldsymbol{x}_t}\log p_t(\boldsymbol{x}_t \mid \boldsymbol{x}_T) - \nabla_{\boldsymbol{x}_t}\log p_t(\boldsymbol{x}_T \mid \boldsymbol{x}_t)\right] \right) \mathrm{d}t \\
&\quad + \sigma_t \sqrt{-2\frac{\mathrm{d}\log\left(\frac{\alpha_t}{\sigma_t}\right)}{\mathrm{d}t}}\,\mathrm{d}\mathbf{w}_t \\
&= \quad \boldsymbol{x}_t \frac{\mathrm{d}\log\alpha_t}{\mathrm{d}t}\mathrm{d}t - 2\sigma_t^2 \frac{\mathrm{d}\log\left(\frac{\alpha_t}{\sigma_t}\right)}{\mathrm{d}t} \left[ \frac{\frac{\alpha_t}{\alpha_T}\boldsymbol{x}_T - \boldsymbol{x}_t}{\sigma_t^2\left(\frac{\alpha_t^2\sigma_T^2}{\sigma_t^2\alpha_T^2} - 1\right)} \right] \mathrm{d}t \\
&\quad + 2\sigma_t^2 \frac{\mathrm{d}\log\left(\frac{\alpha_t}{\sigma_t}\right)}{\mathrm{d}t} \left[ \frac{\frac{\alpha_T^2\sigma_t^2}{\sigma_T^2\alpha_t^2}\frac{\alpha_t}{\alpha_T}\boldsymbol{x}_T + \alpha_t\boldsymbol{D}_{\boldsymbol{\theta}}(\boldsymbol{x}_t)\left(1 - \frac{\alpha_T^2\sigma_t^2}{\sigma_T^2\alpha_t^2}\frac{\alpha_t}{\alpha_T}\right) - \boldsymbol{x}_t}{\sigma_t^2\left(1 - \frac{\alpha_T^2\sigma_t^2}{\sigma_T^2\alpha_t^2}\frac{\alpha_t}{\alpha_T}\right)} \right] \mathrm{d}t \\
&\quad + \sigma_t \sqrt{-2\frac{\mathrm{d}\log\left(\frac{\alpha_t}{\sigma_t}\right)}{\mathrm{d}t}}\,\mathrm{d}\mathbf{w}_t \\
&= \left( \boldsymbol{x}_t \frac{\mathrm{d}\log\alpha_t}{\mathrm{d}t} + 2\sigma_t^2 \frac{\mathrm{d}\log\left(\frac{\alpha_t}{\sigma_t}\right)}{\mathrm{d}t} \left[ \frac{\alpha_t\boldsymbol{D}_{\boldsymbol{\theta}}(\boldsymbol{x}_t) - \boldsymbol{x}_t}{\sigma_t^2} \right] \right) \mathrm{d}t + \sigma_t \sqrt{-2\frac{\mathrm{d}\log\left(\frac{\alpha_t}{\sigma_t}\right)}{\mathrm{d}t}}\,\mathrm{d}\mathbf{w}_t \\
&= \underbrace{\left( \frac{\mathrm{d}\log\alpha_t}{\mathrm{d}t} - 2\frac{\mathrm{d}\log\left(\frac{\alpha_t}{\sigma_t}\right)}{\mathrm{d}t} \right)}_{L(t)\text{ (Linear Term)}} \boldsymbol{x}_t\,\mathrm{d}t \\
&\quad + \underbrace{\left( 2\,\alpha_t \frac{\mathrm{d}\log\left(\frac{\alpha_t}{\sigma_t}\right)}{\mathrm{d}t}\boldsymbol{D}_{\boldsymbol{\theta}}(\boldsymbol{x}_t) + \sigma_t \sqrt{-2\frac{\mathrm{d}\log\left(\frac{\alpha_t}{\sigma_t}\right)}{\mathrm{d}t}}\,\frac{\mathrm{d}\mathbf{w}_t}{\mathrm{d}t} \right)}_{N(\boldsymbol{x}_t, t)\text{ (Non-linear Term)}} \mathrm{d}t,
\end{aligned}
\tag{2}
$$

where $L(t)$ is the *linear* term, and $N(\boldsymbol{x}_t, t)$ is the *non-linear* term. This shows the semi-linearity of Proposition 2 of the main paper. Thanks to the semi-linearity, we can make use of the Exponential Integrators method [3] to solve Equation (2), as explained next.

Given initial value $\boldsymbol{x}_a$, where $b = a + \Delta\tau$ and $0 < b < a < T$, we obtain the solution to $\boldsymbol{x}_b$ as:

$$\boldsymbol{x}_b = e^{\int_a^b L(r)\,\mathrm{d}r}\,\boldsymbol{x}_a + \int_a^b e^{\int_{a+\tau}^b L(h)\,\mathrm{d}h} \cdot N(\boldsymbol{x}_\tau, \tau)\,\mathrm{d}\tau. \tag{3}$$

To simplify the integral of the linear term $L(r)$ from $a$ to $b$, we first expand $L(r)$ using the properties of logarithms:

$$\begin{aligned} L(r) &= \frac{\mathrm{d}\log\alpha_r}{\mathrm{d}r} - 2\frac{\mathrm{d}\log\left(\frac{\alpha_r}{\sigma_r}\right)}{\mathrm{d}r} \\ &= \frac{\mathrm{d}}{\mathrm{d}r}\left[\log\alpha_r - 2(\log\alpha_r - \log\sigma_r)\right] \\ &= \frac{\mathrm{d}}{\mathrm{d}r}\left[2\log\sigma_r - \log\alpha_r\right] \\ &= \frac{\mathrm{d}}{\mathrm{d}r}\left[\log\left(\frac{\sigma_r^2}{\alpha_r}\right)\right]. \end{aligned} \tag{4}$$

Because $L(r)$ simplifies to a single exact derivative, we can easily evaluate its integral with respect to $r$ from $a$ to $b$:

$$\begin{aligned} \int_a^b L(r)\,\mathrm{d}r &= \int_a^b \frac{\mathrm{d}}{\mathrm{d}r}\left[\log\left(\frac{\sigma_r^2}{\alpha_r}\right)\right]\mathrm{d}r \\ &= \left[\log\frac{\sigma_r^2}{\alpha_r}\right]_a^b \\ &= \log\left(\frac{\sigma_b^2}{\alpha_b}\right) - \log\left(\frac{\sigma_a^2}{\alpha_a}\right) \\ &= \log\left(\frac{\alpha_a}{\alpha_b}\cdot\frac{\sigma_b^2}{\sigma_a^2}\right). \end{aligned} \tag{5}$$

Next, let's define $\lambda_t := \log\frac{\alpha_t}{\sigma_t}$, with the inverse function $t_\lambda(\cdot)$, that satisfies $t_\lambda(\lambda_t) = t$. Through the *change-of-variables* method for $\lambda$, we can denote $\alpha_\lambda := \alpha_{t_\lambda(\lambda)}$, $\boldsymbol{x}_\lambda := \boldsymbol{x}_{t_\lambda(\lambda)}$, $\mathbf{D}_\theta(\boldsymbol{x}_\lambda) := \mathbf{D}_\theta(\boldsymbol{x}_{t_\lambda(\lambda)})$, $\mathbf{w}_\lambda := \mathbf{w}_{t_\lambda(\lambda)}$, $\mathrm{d}\mathbf{w}_\lambda := \sqrt{-\frac{\mathrm{d}\lambda}{\mathrm{d}t}}\,\mathrm{d}\mathbf{w}_{t_\lambda(\lambda)}$, and $N(\boldsymbol{x}_\lambda, \lambda) := N(\boldsymbol{x}_{t_\lambda(\lambda)}, t_\lambda(\lambda))$.

Thus, we rewrite Equation (3) as:

$$\begin{aligned}
\boldsymbol{x}_b &= \frac{\alpha_a}{\alpha_b}\frac{\sigma_b^2}{\sigma_a^2}\boldsymbol{x}_a + \int_{\lambda_a}^{\lambda_b} \frac{\alpha_\lambda}{\alpha_b}\frac{\sigma_b^2}{\sigma_\lambda^2} \cdot N(\boldsymbol{x}_\lambda, \lambda)\,\mathrm{d}\lambda \\
&= \frac{\alpha_a}{\alpha_b}\frac{\sigma_b^2}{\sigma_a^2}\boldsymbol{x}_a + \int_{\lambda_a}^{\lambda_b} \frac{\alpha_\lambda}{\alpha_b}\frac{\sigma_b^2}{\sigma_\lambda^2} \left(2\,\alpha_\lambda \boldsymbol{D}_{\boldsymbol{\theta}}(\boldsymbol{x}_\lambda) + \sqrt{2}\,\sigma_\lambda \frac{\mathrm{d}\mathbf{w}_\lambda}{\mathrm{d}\lambda}\right)\mathrm{d}\lambda \\
&= \frac{\alpha_a}{\alpha_b}\frac{\sigma_b^2}{\sigma_a^2}\boldsymbol{x}_a + 2\,\frac{\sigma_b^2}{\alpha_b}\int_{\lambda_a}^{\lambda_b} \frac{\alpha_\lambda^2}{\sigma_\lambda^2}\boldsymbol{D}_{\boldsymbol{\theta}}(\boldsymbol{x}_\lambda)\,\mathrm{d}\lambda + \sqrt{2}\,\frac{\sigma_b^2}{\alpha_b}\int_{\lambda_a}^{\lambda_b} \frac{\alpha_\lambda}{\sigma_\lambda}\mathrm{d}\mathbf{w}_\lambda \\
&= \frac{\alpha_a}{\alpha_b}\frac{\sigma_b^2}{\sigma_a^2}\boldsymbol{x}_a + 2\,\alpha_b\, e^{-2\lambda_b}\underbrace{\int_{\lambda_a}^{\lambda_b} e^{2\lambda}\boldsymbol{D}_{\boldsymbol{\theta}}(\boldsymbol{x}_\lambda)\,\mathrm{d}\lambda}_{\text{Use Taylor Expansion}} + \sqrt{2}\,\alpha_b\, e^{-2\lambda_b}\underbrace{\int_{\lambda_a}^{\lambda_b} e^{\lambda}\,\mathrm{d}\mathbf{w}_\lambda}_{\text{Itô Integral}}\,. \qquad (6)
\end{aligned}$$

The integral $\int_{\lambda_a}^{\lambda_b} e^{2\lambda}\boldsymbol{D}_{\boldsymbol{\theta}}(\boldsymbol{x}_\lambda)\,\mathrm{d}\lambda$ can be computed by performing Taylor Expansion:

$$\int_{\lambda_a}^{\lambda_b} e^{2\lambda}\boldsymbol{D}_{\boldsymbol{\theta}}(\boldsymbol{x}_\lambda)\,\mathrm{d}\lambda \approx \sum_{n=0}^{k-1}\boldsymbol{D}_{\boldsymbol{\theta}}^{(n)}(\boldsymbol{x}_{\lambda_a})\int_{\lambda_a}^{\lambda_b} e^{2\lambda}\frac{(\lambda-\lambda_a)^n}{n!}\,\mathrm{d}\lambda + \mathcal{O}((\lambda_b-\lambda_a)^{k+1}), \qquad (7)$$

where $k \geq 1$, and $\boldsymbol{D}_{\boldsymbol{\theta}}^{(n)}(\boldsymbol{x}_{\lambda_a}) := \frac{\mathrm{d}^n \boldsymbol{D}_{\boldsymbol{\theta}}(\boldsymbol{x}_{\lambda_a})}{\mathrm{d}\lambda^n}$ is the $n^{\text{th}}$-order derivative of $\boldsymbol{D}_{\boldsymbol{\theta}}(\cdot)$ w.r.t. $\lambda$. Furthermore, we can compute the Itô integral [10] as:

$$\int_{\lambda_a}^{\lambda_b} e^{\lambda}\,\mathrm{d}\mathbf{w}_\lambda = \left(\sqrt{\int_{\lambda_a}^{\lambda_b} e^{2\lambda}\,\mathrm{d}\lambda}\right)\mathbf{z}_b = \frac{e^{\lambda_b}}{\sqrt{2}}\sqrt{1-e^{2(\lambda_a-\lambda_b)}}\,\mathbf{z}_b,$$

where $\mathbf{z}_b \sim \mathcal{N}(\mathbf{0}, \boldsymbol{I})$.

#### 2.1.1. $1^{\text{st}}$-order Bridge SDE Solution

Substituting $k = 1$ for Equation (7), we can ultimately simplify Equation (6) as:

$$\begin{aligned}
\boldsymbol{x}_b &= \frac{\alpha_b}{\alpha_a}e^{2(\lambda_a-\lambda_b)}\boldsymbol{x}_a + \alpha_b\left(1-e^{2(\lambda_a-\lambda_b)}\right)\boldsymbol{D}_{\boldsymbol{\theta}}(\boldsymbol{x}_{\lambda_a}) + \sigma_b\sqrt{1-e^{2(\lambda_a-\lambda_b)}}\,\mathbf{z}_b \\
&= \frac{\alpha_b}{\alpha_a}\frac{\mathrm{SNR}_a}{\mathrm{SNR}_b}\boldsymbol{x}_a + \alpha_b\left(1-\frac{\mathrm{SNR}_a}{\mathrm{SNR}_b}\right)\boldsymbol{D}_{\boldsymbol{\theta}}(\boldsymbol{x}_{\lambda_a}) + \sigma_b\sqrt{1-\frac{\mathrm{SNR}_a}{\mathrm{SNR}_b}}\,\mathbf{z}_b, \qquad (8)
\end{aligned}$$

where $\mathrm{SNR}_t := \alpha_t^2/\sigma_t^2 = e^{2\lambda_t}$, giving us the $1^{\text{st}}$-order solution to the Bridge SDE. Thus, given the initial value $\boldsymbol{x}_a$, where $b = a + \Delta t$ and $0 \leq b < a \leq T$, the solution to $\boldsymbol{x}_b$ is:

$$\boldsymbol{x}_b = \frac{\alpha_b}{\alpha_a}\frac{\mathrm{SNR}_a}{\mathrm{SNR}_b}\boldsymbol{x}_a + \alpha_b\left(1-\frac{\mathrm{SNR}_a}{\mathrm{SNR}_b}\right)\boldsymbol{D}_{\boldsymbol{\theta}}(\boldsymbol{x}_a) + \sigma_b\sqrt{1-\frac{\mathrm{SNR}_a}{\mathrm{SNR}_b}}\,\mathbf{z}_b. \qquad (9)$$

■

#### 2.1.2. $2^{\text{nd}}$-order Bridge SDE Solution

Similar to the process above, by substituting $k = 2$ for Equation (7), we simplify Equation (6) as:

$$\boldsymbol{x}_b = \frac{\alpha_b}{\alpha_a} e^{2(\lambda_a - \lambda_b)} \boldsymbol{x}_a + \alpha_b \left(1 - e^{2(\lambda_a - \lambda_b)}\right) \boldsymbol{D}_{\boldsymbol{\theta}}(\boldsymbol{x}_{\lambda_a}) \tag{10}$$

$$+ \alpha_b \left(\lambda_b - \lambda_a + e^{2(\lambda_a - \lambda_b)} - 1\right) \boldsymbol{D}_{\boldsymbol{\theta}}^{(1)}(\boldsymbol{x}_{\lambda_a}) + \sigma_b \sqrt{1 - e^{2(\lambda_a - \lambda_b)}}\, \mathbf{z}_b$$

$$= \frac{\text{SNR}_a}{\text{SNR}_b} \frac{\alpha_b}{\alpha_a} \boldsymbol{x}_a + \alpha_b \left(1 - \frac{\text{SNR}_a}{\text{SNR}_b}\right) \boldsymbol{D}_{\boldsymbol{\theta}}(\boldsymbol{x}_{\lambda_a}) \tag{11}$$

$$+ \alpha_b \left(\lambda_b - \lambda_a + \frac{\text{SNR}_a}{\text{SNR}_b} - 1\right) \boldsymbol{D}_{\boldsymbol{\theta}}^{(1)}(\boldsymbol{x}_{\lambda_a}) + \sigma_b \sqrt{1 - \frac{\text{SNR}_a}{\text{SNR}_b}}\, \mathbf{z}_b, \tag{12}$$

where $\boldsymbol{D}_{\boldsymbol{\theta}}^{(n)}(\boldsymbol{x}_{\lambda_a}) := \frac{\mathrm{d}^n \boldsymbol{D}_{\boldsymbol{\theta}}(\boldsymbol{x}_{\lambda_a})}{\mathrm{d}\lambda^n}$ is the $n^{\text{th}}$-order derivative of $\boldsymbol{D}_{\boldsymbol{\theta}}(\cdot)$ w.r.t. $\lambda$, giving us the $2^{\text{nd}}$-order solution to the Bridge SDE.

■

#### 2.1.3. Ablation Study on the Bridge SDE Solutions

We provide an ablation study on the effect of $k = 1$ (Eq. (9)) and $k = 2$ (Eq. (12)). In Section 3 of the main paper, we justify our choice of $k = 1$ for the **initial Bridge SDE step** in DBMSolver. As $k = 2$ step is higher-order, it requires an extra NFE. However, because this initial step is taken over a very small interval $\mathrm{d}t = \epsilon$, $k = 2$'s higher-order solution renders the gains trivial, providing negligible benefits.
To validate this claim, we conduct an ablation study on the DIODE $(256 \times 256)$ dataset using DBMSolver. For a fixed number of sampling steps (with the same $2^{\text{nd}}$-order Bridge ODE step), we compare $k = 1$ and $k = 2$ and report the results in Table 2. For 4 sampling steps, both settings achieve an FID of 3.38. However, $k = 1$ requires only 6 NFEs, whereas $k = 2$ requires 7 NFEs due to its higher-order computation. With 11 sampling steps, the FIDs for $k = 1$ and $k = 2$ are 2.06 and 2.07, respectively. Although the IS metric for $k = 2$ is slightly better (6.04 IS) compared to that of $k = 1$ (6.00 IS), the improvement is marginal compared to the computational cost incurred for the extra NFE .
These results confirm that using $k = 2$ offers no meaningful improvement in generation quality over $k = 1$, supporting our choice to use $k = 1$ for the initial Bridge SDE step.

Table 2. Ablation Study on value of $k$ for the Bridge SDE solution on DIODE $(256 \times 256)$. We see that compared to using $k = 2$, using $k = 1$ for the solution for DBMSolver is sufficiently good and requires lower computational resources.

| Value of $k$ | Steps | NFEs | FID ↓ | IS ↑ | LPIPS ↓ | MSE ↓ |
|---|---|---|---|---|---|---|
| $k = 2$ | 4 | 7 | 3.38 | **6.04** | 0.197 | **0.015** |
| $k = 2$ | 11 | 21 | 2.07 | 6.02 | 0.198 | 0.018 |
| **$k = 1$ (Proposed)** | 4 | 6 | 3.38 | 6.00 | **0.196** | **0.015** |
| **$k = 1$ (Proposed)** | 11 | 20 | **2.06** | 6.00 | 0.198 | 0.018 |

### 2.2. Proof of Proposition 2 of the main paper

Given a well-trained DBM $\boldsymbol{D}_\theta(\cdot)$ that approximates data sample $\boldsymbol{x}_0$, we can re-write the Bridge PF ODE (Equation 5 of the main paper) as:

$$
\begin{aligned}
\frac{\mathrm{d}\boldsymbol{x}_t}{\mathrm{d}t} &= f(\boldsymbol{x}_t, t) - g(t)^2 \left( \frac{1}{2} \nabla_{\boldsymbol{x}_t} \log p_t(\boldsymbol{x}_t \mid \boldsymbol{x}_T) - \nabla_{\boldsymbol{x}_t} \log p_t(\boldsymbol{x}_T \mid \boldsymbol{x}_t) \right) \\
&= \boldsymbol{x}_t \frac{\mathrm{d} \log \alpha_t}{\mathrm{d}t} + 2\sigma_t^2 \frac{\mathrm{d} \log \left( \frac{\alpha_t}{\sigma_t} \right)}{\mathrm{d}t} \left( \frac{1}{2} \nabla_{\boldsymbol{x}_t} \log p_t(\boldsymbol{x}_t \mid \boldsymbol{x}_T) - \nabla_{\boldsymbol{x}_t} \log p_t(\boldsymbol{x}_T \mid \boldsymbol{x}_t) \right) \\
&= \boldsymbol{x}_t \frac{\mathrm{d} \log \alpha_t}{\mathrm{d}t} + \sigma_t^2 \frac{\mathrm{d} \log \left( \frac{\alpha_t}{\sigma_t} \right)}{\mathrm{d}t} \cdot \frac{\frac{\alpha_T^2 \sigma_t^2}{\sigma_T^2 \alpha_t^2} \frac{\alpha_t}{\alpha_T} \boldsymbol{x}_T + \alpha_t \left( 1 - \frac{\alpha_T^2 \sigma_t^2}{\sigma_T^2 \alpha_t^2} \right) \boldsymbol{D}_{\boldsymbol{\theta}}(\boldsymbol{x}_t) - \boldsymbol{x}_t}{\sigma_t^2 \left( 1 - \frac{\alpha_T^2 \sigma_t^2}{\sigma_T^2 \alpha_t^2} \right)} \\
&\quad - 2\sigma_t^2 \frac{\mathrm{d} \log \left( \frac{\alpha_t}{\sigma_t} \right)}{\mathrm{d}t} \cdot \frac{\frac{\alpha_t}{\alpha_T} \boldsymbol{x}_T - \boldsymbol{x}_t}{\sigma_t^2 \left( \frac{\alpha_t^2 \sigma_T^2}{\sigma_t^2 \alpha_T^2} - 1 \right)} \\
&= \boldsymbol{x}_t \frac{\mathrm{d} \log \alpha_t}{\mathrm{d}t} + \frac{\mathrm{d} \log \left( \frac{\alpha_t}{\sigma_t} \right)}{\mathrm{d}t} \left[ \alpha_t \boldsymbol{D}_{\boldsymbol{\theta}}(\boldsymbol{x}_t) - \boldsymbol{x}_t - \frac{\frac{\alpha_t}{\alpha_T} \boldsymbol{x}_T - \boldsymbol{x}_t}{\frac{\alpha_t^2 \sigma_T^2}{\sigma_t^2 \alpha_T^2} - 1} \right] \\
&= \boldsymbol{x}_t \left( \frac{\mathrm{d} \log \alpha_t}{\mathrm{d}t} - \frac{\mathrm{d} \log \left( \frac{\alpha_t}{\sigma_t} \right)}{\mathrm{d}t} + \frac{1}{\frac{\alpha_t^2 \sigma_T^2}{\sigma_t^2 \alpha_T^2} - 1} \frac{\mathrm{d} \log \left( \frac{\alpha_t}{\sigma_t} \right)}{\mathrm{d}t} \right) \\
&\quad + \alpha_t \frac{\mathrm{d} \log \left( \frac{\alpha_t}{\sigma_t} \right)}{\mathrm{d}t} \left[ \boldsymbol{D}_{\boldsymbol{\theta}}(\boldsymbol{x}_t) - \frac{\boldsymbol{x}_T / \alpha_T}{\frac{\alpha_t^2 \sigma_T^2}{\sigma_t^2 \alpha_T^2} - 1} \right].
\end{aligned}
$$

We further simplify the equation above as:

$$\frac{\mathrm{d}\boldsymbol{x}_t}{\mathrm{d}t} = \underbrace{\boldsymbol{x}_t \left( \frac{\mathrm{d}\log\sigma_t}{\mathrm{d}t} + \frac{1}{\frac{\alpha_t^2\sigma_T^2}{\sigma_t^2\alpha_T^2} - 1} \frac{\mathrm{d}\log\left(\frac{\alpha_t}{\sigma_t}\right)}{\mathrm{d}t} \right)}_{L(t)\text{ (Linear Term)}} + \underbrace{\alpha_t \left[ \boldsymbol{D}_{\boldsymbol{\theta}}(\boldsymbol{x}_t) - \frac{\boldsymbol{x}_T/\alpha_T}{\frac{\alpha_t^2\sigma_T^2}{\sigma_t^2\alpha_T^2} - 1} \right] \frac{\mathrm{d}\log\left(\frac{\alpha_t}{\sigma_t}\right)}{\mathrm{d}t}}_{N(\boldsymbol{x}_t, t)\text{ (Non-linear Term)}}. \tag{13}$$

where $L(t)$ is the *linear* term, and $N(\boldsymbol{x}_t, t)$ is the *non-linear* term. Thus, we can clearly observe the semi-linearity of Equation 5 of the main paper. Similar to the derivation above, we can once again make use of the EI method to solve Equation (13):
Given an initial value $\boldsymbol{x}_a$ where $b = a + \Delta\tau$ and $0 < b < a < T$, we obtain the solution to $\boldsymbol{x}_b$ as:

$$\boldsymbol{x}_b = e^{\int_a^b L(r)\,\mathrm{d}r}\,\boldsymbol{x}_a + \int_a^b e^{\int_{a+\tau}^b L(h)\,\mathrm{d}h} \cdot N(\boldsymbol{x}_\tau, \tau)\,\mathrm{d}\tau. \tag{14}$$

To simplify the equation, we first integrate the linear term $L(r)$ with respect to $r$ from $a$ to $b$:

$$\int_a^b L(r)\,\mathrm{d}r = \left[ \log\left( \frac{\alpha_r\sqrt{e^{2(\lambda_r-\lambda_T)} - 1}}{e^{2\lambda_r}} \right) \right]_a^b = \log\left( \frac{\alpha_b}{\alpha_a} e^{2(\lambda_a-\lambda_b)} \sqrt{\frac{e^{2(\lambda_b-\lambda_T)} - 1}{e^{2(\lambda_a-\lambda_T)} - 1}} \right), \tag{15}$$

where $\lambda_t := \log\frac{\alpha_t}{\sigma_t}$ has the inverse function $t_\lambda(\cdot)$ which satisfies $t_\lambda(\lambda_t) = t$.
Next, we can use the *change-of-variables* method for $\lambda$. We denote $\alpha_\lambda := \alpha_{t_\lambda(\lambda)}$, $\boldsymbol{x}_\lambda := \boldsymbol{x}_{t_\lambda(\lambda)}$, and $N(\boldsymbol{x}_\lambda, \lambda) := N(\boldsymbol{x}_{t_\lambda(\lambda)}, t_\lambda(\lambda))$.
Substituting Equation (15) along with the value of $N(\boldsymbol{x}_\lambda, \lambda)$ back into Equation (14), we get:

$$\begin{aligned}\boldsymbol{x}_b = &\frac{\alpha_b}{\alpha_a} e^{2(\lambda_a-\lambda_b)} \sqrt{\frac{e^{2(\lambda_b-\lambda_T)} - 1}{e^{2(\lambda_a-\lambda_T)} - 1}}\,\boldsymbol{x}_a \\ &+ \alpha_b\,e^{-2\lambda_b} \underbrace{\int_{\lambda_a}^{\lambda_b} \frac{e^{2\lambda}}{\alpha_\lambda} \sqrt{\frac{e^{2(\lambda_b-\lambda_T)} - 1}{e^{2(\lambda-\lambda_T)} - 1}} \cdot \alpha_\lambda \left[ \boldsymbol{D}_{\boldsymbol{\theta}}(\boldsymbol{x}_\lambda) - \frac{\boldsymbol{x}_T/\alpha_T}{e^{2(\lambda-\lambda_T)} - 1} \right] \mathrm{d}\lambda}_{\text{Separate and simplify integral further}}\end{aligned} \tag{16}$$

When we simplify the integral in Equation (16), we get:

$$\sqrt{e^{2(\lambda_b-\lambda_T)}-1}\int_{\lambda_a}^{\lambda_b}\left[\frac{e^{2\lambda}\boldsymbol{D}_{\boldsymbol{\theta}}(\boldsymbol{x}_\lambda)}{\sqrt{e^{2(\lambda-\lambda_T)}-1}}-\frac{\boldsymbol{x}_T\,e^{2\lambda}}{\alpha_T\,\left(e^{2(\lambda-\lambda_T)}-1\right)^{3/2}}\right]\mathrm{d}\lambda. \tag{17}$$

Next, we separately integrate the individual terms of Equation (17).
For the integral associated with $\boldsymbol{x}_T$, we use *integration-by-parts* to simplify it:

$$\int_{\lambda_a}^{\lambda_b}\frac{\boldsymbol{x}_T\,e^{2\lambda}}{\alpha_T\,\left(e^{2(\lambda-\lambda_T)}-1\right)^{3/2}}\mathrm{d}\lambda=-\frac{\boldsymbol{x}_T}{\alpha_T}\frac{e^{2\lambda_T}}{\sqrt{e^{2(\lambda_b-\lambda_T)}-1}}\left(1-\sqrt{\frac{e^{2(\lambda_b-\lambda_T)}-1}{e^{2(\lambda_a-\lambda_T)}-1}}\right) \tag{18}$$

Using the result in Equation (18), we can simplify Equation (16) as:

$$\begin{aligned}\boldsymbol{x}_b=&\frac{\alpha_b}{\alpha_a}e^{2(\lambda_a-\lambda_b)}\sqrt{\frac{\rho(\lambda_b,\lambda_T)}{\rho(\lambda_a,\lambda_T)}}\,\boldsymbol{x}_a+\frac{\alpha_b}{\alpha_T}e^{2(\lambda_T-\lambda_b)}\left(1-\sqrt{\frac{\rho(\lambda_b,\lambda_T)}{\rho(\lambda_a,\lambda_T)}}\right)\boldsymbol{x}_T\\&+\alpha_b\,e^{-2\lambda_b}\sqrt{\rho(\lambda_b,\lambda_T)}\underbrace{\int_{\lambda_a}^{\lambda_b}\frac{e^{2\lambda}\boldsymbol{D}_{\boldsymbol{\theta}}(\boldsymbol{x}_\lambda)}{\sqrt{\rho(\lambda,\lambda_T)}}\mathrm{d}\lambda}_{\text{Use Taylor Expansion}},\end{aligned} \tag{19}$$

where $\rho(m,n):=e^{2(m-n)}-1$.
Finally, we perform Taylor Expansion to obtain the solution to the integral in Equation (19):

$$\int_{\lambda_a}^{\lambda_b}\frac{e^{2\lambda}\boldsymbol{D}_{\boldsymbol{\theta}}(\boldsymbol{x}_\lambda)}{\sqrt{\rho(\lambda,\lambda_T)}}\mathrm{d}\lambda\approx\sum_{n=0}^{k-1}\underbrace{\boldsymbol{D}_{\boldsymbol{\theta}}^{(n)}(\boldsymbol{x}_{\lambda_a})}_{\text{Estimated}}\underbrace{\int_{\lambda_a}^{\lambda_b}\frac{e^{2\lambda}}{\sqrt{\rho(\lambda,\lambda_T)}}\frac{(\lambda-\lambda_a)^n}{n!}\mathrm{d}\lambda}_{\text{Analytically Computed (Section 2.3)}}+\underbrace{\mathcal{O}((\lambda_b-\lambda_a)^{k+1})}_{\text{Omitted}},$$

where $k\geq 1$, and $\boldsymbol{D}_{\boldsymbol{\theta}}^{(n)}(\boldsymbol{x}_{\lambda_a}):=\frac{\mathrm{d}^n\boldsymbol{D}_{\boldsymbol{\theta}}(\boldsymbol{x}_{\lambda_a})}{\mathrm{d}\lambda^n}$ is the $n^{\text{th}}$-order derivative of $\boldsymbol{D}_{\boldsymbol{\theta}}(\cdot)$ w.r.t. $\lambda$.
This is the same as Equation 10 of the main paper.
Thus, we can derive an exact solution for $\boldsymbol{x}_b$. For completeness, we derive the $1^{\text{st}}$- and $2^{\text{nd}}$-order solutions below.

■

### 2.3. Deriving Solutions for Proposition 2 of the main paper

#### 2.3.1. 1st-order Bridge ODE Solution

We use Taylor Expansion to find the solution. By using $k = 1$, the 1st-order solution is as follows:

$$\begin{aligned}\int_{\lambda_a}^{\lambda_b} \frac{e^{2\lambda}\boldsymbol{D}_{\boldsymbol{\theta}}(\boldsymbol{x}_\lambda)}{\sqrt{\rho(\lambda,\lambda_T)}}\,\mathrm{d}\lambda &\approx \boldsymbol{D}_{\boldsymbol{\theta}}^{(0)}(\boldsymbol{x}_{\lambda_a})\int_{\lambda_a}^{\lambda_b} \frac{e^{2\lambda}}{\sqrt{\rho(\lambda,\lambda_T)}}\frac{(\lambda-\lambda_a)^0}{0!}\mathrm{d}\lambda \\ &= \boldsymbol{D}_{\boldsymbol{\theta}}(\boldsymbol{x}_{\lambda_a})\int_{\lambda_a}^{\lambda_b} \frac{e^{2\lambda}}{\sqrt{\rho(\lambda,\lambda_T)}}\,\mathrm{d}\lambda \\ &= \boldsymbol{D}_{\boldsymbol{\theta}}(\boldsymbol{x}_{\lambda_a})\,e^{2\lambda_T}\sqrt{\rho(\lambda_b,\lambda_T)}\left(1-\sqrt{\frac{\rho(\lambda_a,\lambda_T)}{\rho(\lambda_b,\lambda_T)}}\right). \end{aligned} \tag{20}$$

Substituting Equation (20) back into Equation (19), we get the following 1st-order formulation of $\boldsymbol{x}_b$:

$$\begin{aligned}\boldsymbol{x}_b = \frac{\alpha_b}{\alpha_a}e^{2(\lambda_a-\lambda_b)}\sqrt{\frac{\rho(\lambda_b,\lambda_T)}{\rho(\lambda_a,\lambda_T)}}\,\boldsymbol{x}_a + \frac{\alpha_b}{\alpha_T}e^{2(\lambda_T-\lambda_b)}\left(1-\sqrt{\frac{\rho(\lambda_b,\lambda_T)}{\rho(\lambda_a,\lambda_T)}}\right)\boldsymbol{x}_T \\ + \alpha_b\,e^{2(\lambda_T-\lambda_b)}\rho(\lambda_b,\lambda_T)\left(1-\sqrt{\frac{\rho(\lambda_a,\lambda_T)}{\rho(\lambda_b,\lambda_T)}}\right)\boldsymbol{D}_{\boldsymbol{\theta}}(\boldsymbol{x}_{\lambda_a}). \end{aligned} \tag{21}$$

■

**Relation to DBIM Sampler [14].** Our solution is a generalized form of the DBIM-1 sampler's deterministic step. Simplifying the DBIM-1 deterministic step's formulation for $\boldsymbol{x}_b$

(with $\rho = 0$), we get:

$$\begin{aligned}
\boldsymbol{x}_b &= \frac{\alpha_b}{\alpha_T}\frac{\text{SNR}_T}{\text{SNR}_b}\boldsymbol{x}_T + \alpha_b\left(1 - \frac{\text{SNR}_T}{\text{SNR}_b}\right)\boldsymbol{D}_\theta(\boldsymbol{x}_a) \\
&\quad + \sigma_b\sqrt{1 - \frac{\text{SNR}_T}{\text{SNR}_b}}\left[\frac{\boldsymbol{x}_a - \frac{\alpha_a}{\alpha_T}\frac{\text{SNR}_T}{\text{SNR}_a}\boldsymbol{x}_T - \alpha_a\left(1 - \frac{\text{SNR}_T}{\text{SNR}_a}\right)\boldsymbol{D}_\theta(\boldsymbol{x}_a)}{\sigma_a\sqrt{1 - \frac{\text{SNR}_T}{\text{SNR}_a}}}\right] \\
&= \frac{\alpha_b}{\alpha_a}\frac{\text{SNR}_a}{\text{SNR}_b}\sqrt{\frac{\frac{\text{SNR}_b}{\text{SNR}_T} - 1}{\frac{\text{SNR}_a}{\text{SNR}_T} - 1}}\boldsymbol{x}_a + \frac{\alpha_b}{\alpha_T}\frac{\text{SNR}_T}{\text{SNR}_b}\left(1 - \frac{\sigma_b\sqrt{1 - \frac{\text{SNR}_T}{\text{SNR}_b}}}{\sigma_a\sqrt{1 - \frac{\text{SNR}_T}{\text{SNR}_a}}}\frac{\alpha_a}{\alpha_T}\frac{\text{SNR}_T}{\text{SNR}_a}\frac{\alpha_T}{\alpha_b}\frac{\text{SNR}_b}{\text{SNR}_T}\right)\boldsymbol{x}_T \\
&\quad + \alpha_b\frac{\text{SNR}_T}{\text{SNR}_b}\left(\frac{\text{SNR}_b}{\text{SNR}_T} - 1\right)\left(1 - \sqrt{\frac{\frac{\text{SNR}_a}{\text{SNR}_T} - 1}{\frac{\text{SNR}_b}{\text{SNR}_T} - 1}}\right)\boldsymbol{D}_\theta(\boldsymbol{x}_a). \qquad (22)
\end{aligned}$$

By substituting the equation above with $e^{2\lambda_t} := \text{SNR}_t$, we in fact see that Equation (22) simplifies to our 1st-order formulation of $\boldsymbol{x}_b$ in Equation (21). Thus, we see that DBIM is actually a 1st-order formulation of our solution to the Bridge PF ODE (*i.e.*, $k = 1$). DBMSolver's advantage is that it instead utilizes a more precise, 2nd-order solution that has lower error bounds compared to DBIM.

#### 2.3.2. 2nd-order Bridge ODE Solution

Similar to the derivation in Section 2.3.1, we use Taylor Expansion to find the solution when $k = 2$:

$$\int_{\lambda_a}^{\lambda_b} \frac{e^{2\lambda}\boldsymbol{D}_{\boldsymbol{\theta}}(\boldsymbol{x}_\lambda)}{\sqrt{\rho(\lambda,\lambda_T)}} \approx \underbrace{\boldsymbol{D}_{\boldsymbol{\theta}}^{(0)}(\boldsymbol{x}_{\lambda_a})\int_{\lambda_a}^{\lambda_b}\frac{e^{2\lambda}}{\sqrt{\rho(\lambda,\lambda_T)}}\,\mathrm{d}\lambda}_{\text{Solution derived in Equation (20)}} + \boldsymbol{D}_{\boldsymbol{\theta}}^{(1)}(\boldsymbol{x}_{\lambda_a})\underbrace{\int_{\lambda_a}^{\lambda_b}\frac{e^{2\lambda}(\lambda-\lambda_a)}{\sqrt{\rho(\lambda,\lambda_T)}}\,\mathrm{d}\lambda}_{\text{Solution derived below}}. \qquad (23)$$

The second term's integral can be solved as:

$$\begin{aligned}
\int_{\lambda_a}^{\lambda_b}\frac{e^{2\lambda}(\lambda-\lambda_a)}{\sqrt{\rho(\lambda,\lambda_T)}}\,\mathrm{d}\lambda &= e^{2\lambda_T}\left[\tan^{-1}\left(\sqrt{\rho(\lambda_b,\lambda_T)}\right) - \tan^{-1}\left(\sqrt{\rho(\lambda_a,\lambda_T)}\right)\right] \\
&\quad + e^{2\lambda_T}\left[(\lambda_b - \lambda_a - 1)\sqrt{\rho(\lambda_b,\lambda_T)} + \sqrt{\rho(\lambda_a,\lambda_T)}\right]. \qquad (24)
\end{aligned}$$

By substituting Equations 20 and 24 into Equation (23), we reach the 2nd-order Taylor

Expansion:

$$e^{2\lambda_T}\sqrt{\rho(\lambda_b,\lambda_T)}\left(1-\sqrt{\frac{\rho(\lambda_a,\lambda_T)}{\rho(\lambda_b,\lambda_T)}}\right)\left[\boldsymbol{D}_{\boldsymbol{\theta}}(\boldsymbol{x}_{\lambda_a})-\boldsymbol{D}_{\boldsymbol{\theta}}^{(1)}(\boldsymbol{x}_{\lambda_a})\right]$$
$$+e^{2\lambda_T}\sqrt{\rho(\lambda_b,\lambda_T)}\left[\lambda_b-\lambda_a+\frac{\tan^{-1}\left(\sqrt{\rho(\lambda_b,\lambda_T)}\right)-\tan^{-1}\left(\sqrt{\rho(\lambda_a,\lambda_T)}\right)}{\sqrt{\rho(\lambda_b,\lambda_T)}}\right]\boldsymbol{D}_{\boldsymbol{\theta}}^{(1)}(\boldsymbol{x}_{\lambda_a}). \tag{25}$$

Finally, substituting Equation (25) back into Equation (19), we get the following formulation of $\boldsymbol{x}_b$:

$$\boldsymbol{x}_b=\frac{\alpha_b}{\alpha_a}e^{2(\lambda_a-\lambda_b)}\sqrt{\frac{\rho(\lambda_b,\lambda_T)}{\rho(\lambda_a,\lambda_T)}}\boldsymbol{x}_a+\frac{\alpha_b}{\alpha_T}e^{2(\lambda_T-\lambda_b)}\left(1-\sqrt{\frac{\rho(\lambda_b,\lambda_T)}{\rho(\lambda_a,\lambda_T)}}\right)\boldsymbol{x}_T$$
$$-\alpha_b\,\rho(\lambda_T,\lambda_b)\left[1-\sqrt{\frac{\rho(\lambda_a,\lambda_T)}{\rho(\lambda_b,\lambda_T)}}\right]\left(\boldsymbol{D}_{\boldsymbol{\theta}}(\boldsymbol{x}_{\lambda_a})-\boldsymbol{D}_{\boldsymbol{\theta}}^{(1)}(\boldsymbol{x}_{\lambda_a})\right)$$
$$-\alpha_b\,\rho(\lambda_T,\lambda_b)\left[\lambda_b-\lambda_a+\frac{\tan^{-1}\left(\sqrt{\rho(\lambda_b,\lambda_T)}\right)-\tan^{-1}\left(\sqrt{\rho(\lambda_a,\lambda_T)}\right)}{\sqrt{\rho(\lambda_b,\lambda_T)}}\right]\boldsymbol{D}_{\boldsymbol{\theta}}^{(1)}(\boldsymbol{x}_{\lambda_a}), \tag{26}$$

where $\boldsymbol{D}_{\boldsymbol{\theta}}^{(n)}(\boldsymbol{x}_{\lambda_a}) := \frac{\mathrm{d}^n\boldsymbol{D}_{\boldsymbol{\theta}}(\boldsymbol{x}_{\lambda_a})}{\mathrm{d}\lambda^n}$ is the $n^{\text{th}}$-order derivative of $\boldsymbol{D}_{\boldsymbol{\theta}}(\cdot)$ w.r.t. $\lambda$. Note that $\tilde{\boldsymbol{x}}_{\lambda_m}$, the perturbed image at $\lambda_m$, is obtained via Equation (21).

■

## 3. Experiment Details

### 3.1. Training Details

We provide thorough details for the diffusion bridge models and their training procedures in Table 3.

### 3.2. Sampling Details

To ensure consistency across different Image-to-Image Translation tasks, the sampling hyperparameters used in our experiments are summarized in Table 4. The following table outlines the discretization schedules and step counts applied to each dataset.

Table 3. Training details for the various Image-to-Image Translation tasks.

| Dataset | Edges2Handbags [5] | DIODE [13] | Inpainting on Conditional ImageNet [2] | CelebAMask-HQ [7] | Face2Comics [12] |
|---|---|---|---|---|---|
| **Hyperparameters and Training Details** | | | | | |
| Bridge Formulation | VP | VP | $I^2$SB [8] | VP | VP |
| Noise Conditioning, $c_{\text{noise}}$ | $250 \ln t$ | $250 \ln t$ | $1000\, t$ | $250 \ln t$ | $250 \ln t$ |
| Learning Rate | 1e-4 | 1e-4 | 1e-4 | 1e-4 | 1e-4 |
| EMA Rate | 0.9999 | 0.9999 | 0.9999 | 0.9993 | 0.9993 |
| Noise Discretization Schedule | Karras | Karras | Karras | Karras | Karras |
| Noise Discretization Steps | 40 | 40 | 40 | 40 | 40 |
| Batch Size | 256 | 64 | 256 | 64 | 64 |
| Training Iterations | 400k | 400k | 400k | 120k | 120k |
| Number and Type of GPUs | 4 A100 | 4 A100 | 8 A800 | 8 A6000 | 8 A6000 |
| **Model Details** | | | | | |
| Model Channels | 192 | 256 | 256 | 256 | 256 |
| Dropout | 10% | 10% | 10% | 10% | 10% |
| Time Embedding | Cosine | Cosine | Cosine | Cosine | Cosine |
| Channel Multiplier | (1, 2, 3, 4) | (1, 1, 2, 2, 4, 4) | (1, 1, 2, 2, 4, 4) | (1, 1, 2, 2, 4, 4) | (1, 1, 2, 2, 4, 4) |
| Number of Residual Layers | 3 | 2 | 2 | 2 | 2 |
| Attention Resolutions | (8, 16, 32) | (8, 16, 32) | (8, 16, 32) | (8, 16, 32) | (8, 16, 32) |
| Model Capacity (Mparams) | 284 | 534 | 534 | 534 | 534 |

Table 4. Sampling details for the various Image-to-Image Translation tasks.

| Dataset | Edges2Handbags [5] | DIODE [13] | Inpainting on Conditional ImageNet [2] | CelebAMask-HQ [7] | Face2Comics [12] |
|---|---|---|---|---|---|
| **Hyperparameters for Sampling** | | | | | |
| Discretization Schedule | Karras | Karras | Uniform | Uniform | Karras |
| Discretization Steps | 40 | 40 | 40 | 40 | 40 |

## 4. More Qualitative Results

Beyond the quantitative metrics presented earlier, we include additional qualitative results in Figures 1–3 to further highlight the perceptual advantages of DBMSolver. We generate images with different random seeds to observe the image generation quality across multiple instances, as portrayed in Figure 1. We observe that different random seeds lead to the generation of different images. This is because the random seed affects the initial Bridge SDE step of DBMSolver, which utilizes random gaussian noise.

In Figures 2 and 3, we continue to observe that our DBMSolver produces significantly better qualitative results compared to other baselines. DBMSolver achieves 34.76 FID in 6 NFEs, while DBIM-1 achieves 44.92 FID on CelebAMask-HQ $(256 \times 256)$. On the Class-conditional Inpainting task on ImageNet $(256 \times 256)$, we achieve 4.98 FID in 6 NFEs while DBIM-1 achieves 5.36 FID, demonstrating our DBMSolver's improvements.

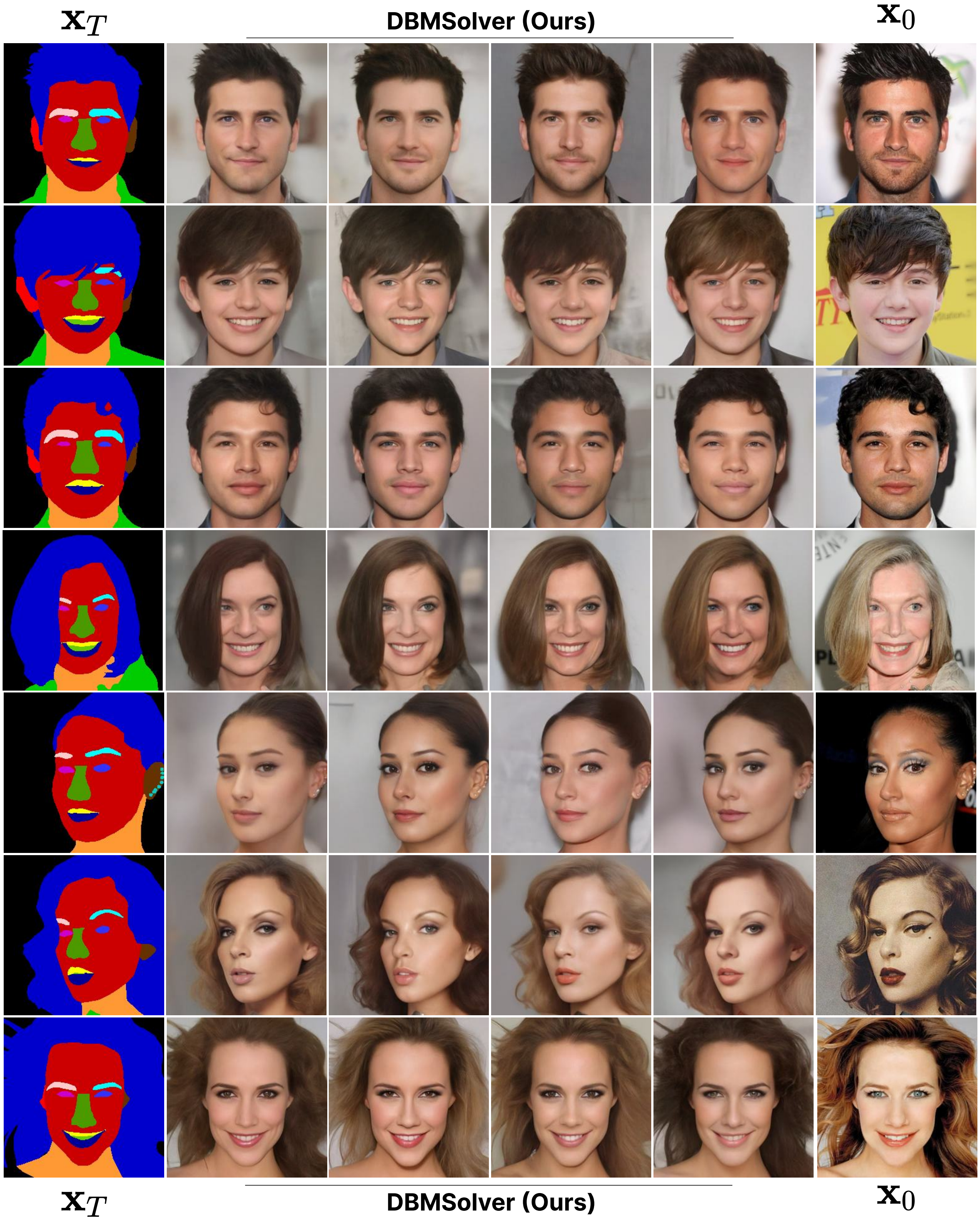


Figure 1. Additional CelebAMask-HQ samples for DBMSolver with 6 NFEs, with different initial SDE steps.

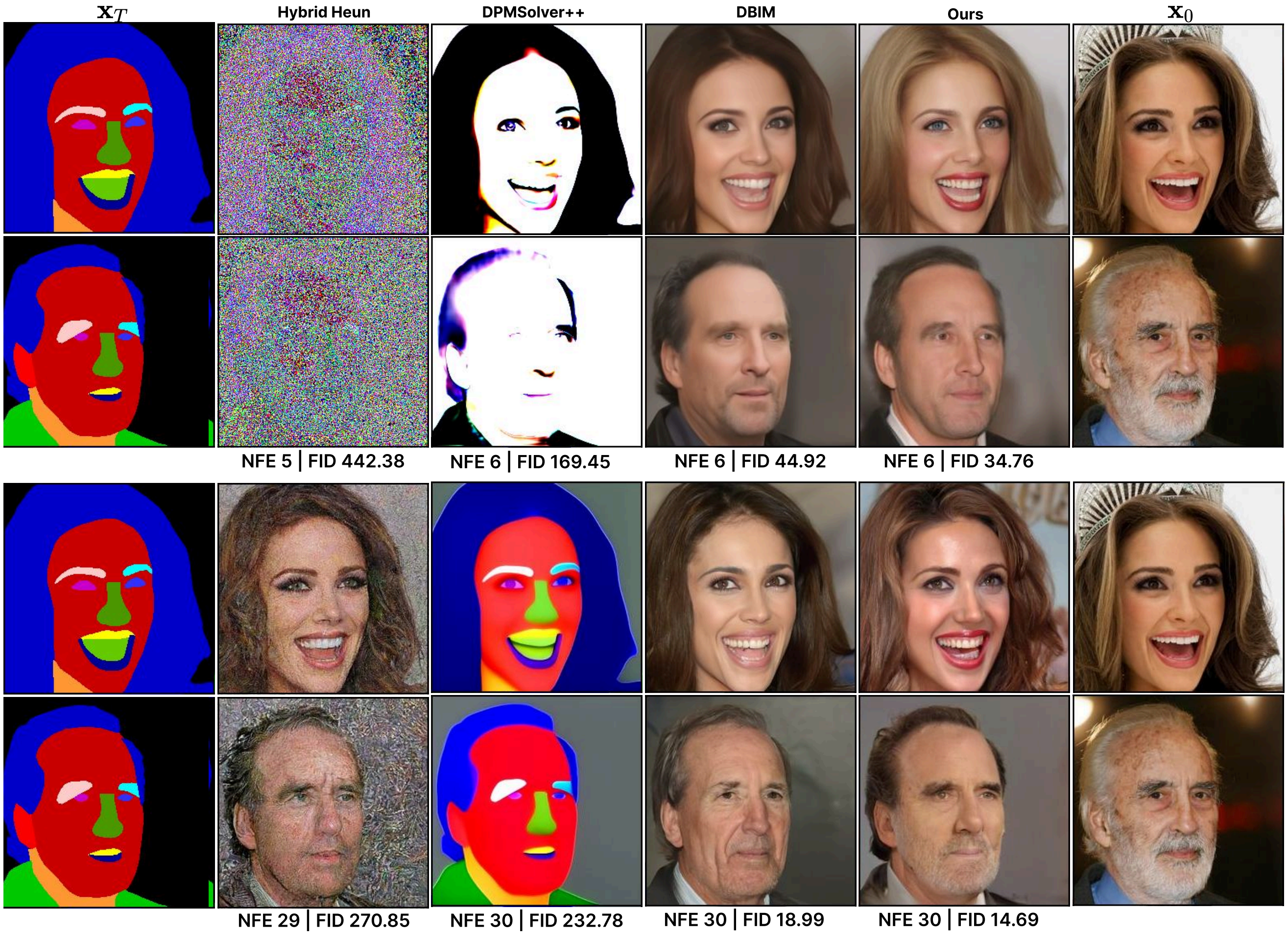


Figure 2. Additional qualitative comparison for Label-to-Face Generation on CelebAMask-HQ.

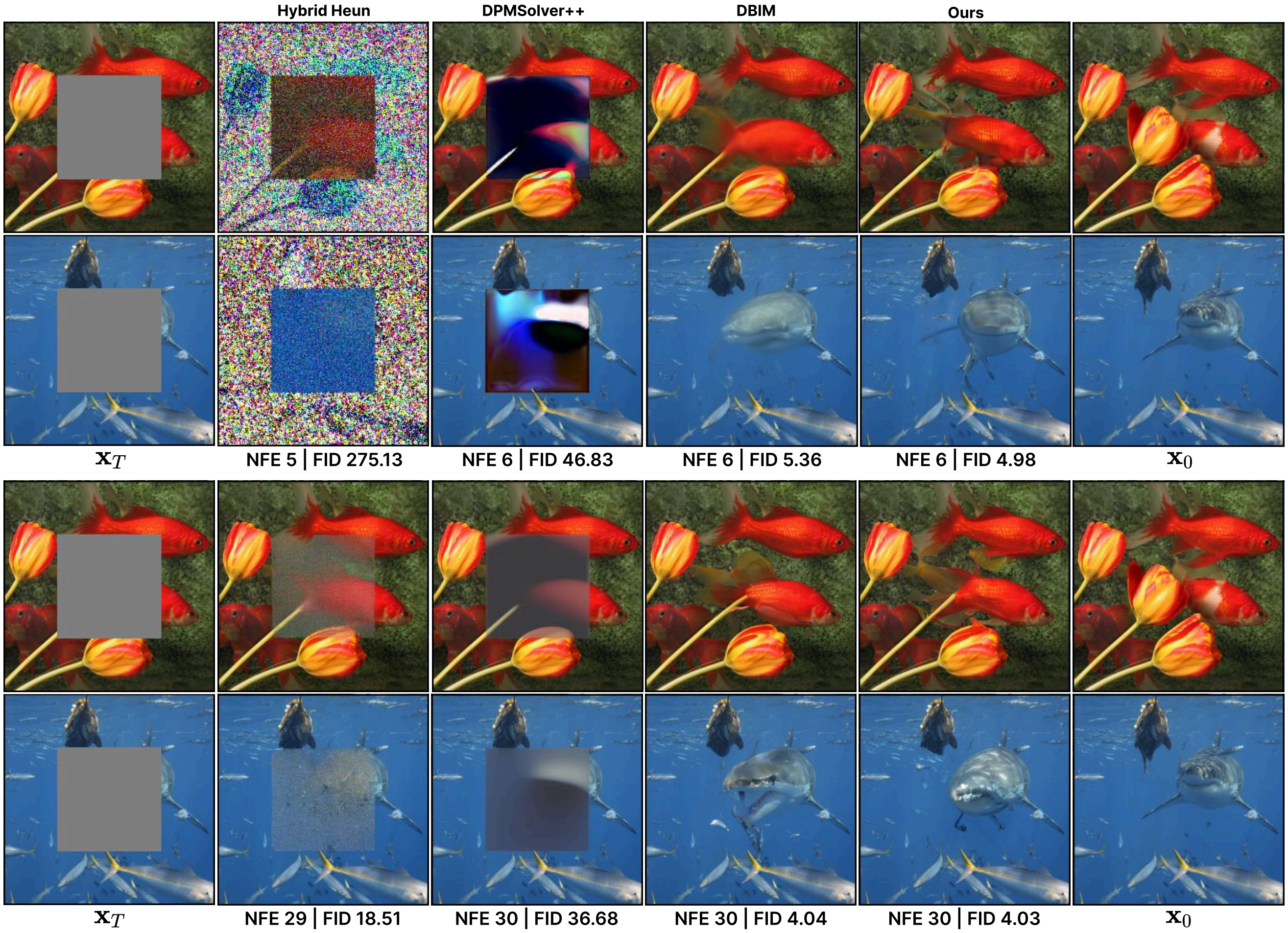


Figure 3. Additional qualitative comparison for Class-Conditional Inpainting on ImageNet.



\end{document}